\documentclass[11pt]{article}

\usepackage[final]{acl}
\usepackage{amsmath}
\usepackage{enumitem}
\usepackage{times}
\usepackage{latexsym}
\usepackage{listings}
\usepackage{minitoc}
\usepackage{varwidth}
\mtcsettitle{minitoc}{Appendix} %
\usepackage[T1]{fontenc}

\usepackage[utf8]{inputenc}
\usepackage{booktabs}  
\usepackage{textcomp}
\usepackage{multirow}   
\usepackage{amssymb}    
\usepackage{mathtools}
\usepackage{graphicx}  
\usepackage{microtype}
\usepackage{tcolorbox}
\tcbuselibrary{breakable}
\usepackage{titletoc} 
\usepackage{inconsolata}
\usepackage{colortbl}
\usepackage{graphicx}
\usepackage{pifont}
\usepackage{booktabs}  
\usepackage{multirow}  
\usepackage{makecell}  
\usepackage{graphicx}  
\usepackage{pifont}    
\usepackage[table]{xcolor} 

\usepackage{amssymb}       

\definecolor{darkgreen}{HTML}{2E8B57}
\definecolor{darkred}{HTML}{D1191F}
\definecolor{myyellow}{HTML}{DAA520}  
\definecolor{graybg}{gray}{0.86}         
\definecolor{naivebg}{gray}{0.95}      
\definecolor{experiencebg}{HTML}{EDF3FF} 
\definecolor{capabilitybg}{HTML}{EAF4E6} 
\definecolor{oursbg}{HTML}{F2EAF8}       

\newcommand{\yes}{\textcolor{darkgreen}{\ding{51}}}
\newcommand{\no}{\textcolor{darkred}{\ding{55}}}
\newcommand{\half}{\textcolor{myyellow}{\ding{52}\rotatebox[origin=c]{-9.2}{\kern-0.7em\ding{55}}}}

\newcommand{\appsection}[1]{%
    \subsection[#1]{#1 \hfill \hyperlink{appendixtoc}{\footnotesize\fbox{Back to ToC}}}%
}

\definecolor{codeblue}{rgb}{0.13,0.29,0.53}  
\definecolor{codegreen}{rgb}{0,0.6,0}        
\definecolor{codegray}{rgb}{0.5,0.5,0.5}     
\definecolor{codepurple}{rgb}{0.58,0,0.82}   
\definecolor{codestring}{rgb}{0.64,0.08,0.08} 
\definecolor{backcolour}{rgb}{0.95,0.95,0.92}

\lstdefinestyle{pythonstyle}{
    language=Python,
    commentstyle=\color{codegreen},
    keywordstyle=\color{codeblue}\bfseries, 
    numberstyle=\tiny\color{codegray},
    stringstyle=\color{codestring},       
    basicstyle=\ttfamily\small,
    breakatwhitespace=false,
    breaklines=true,
    captionpos=b,
    keepspaces=true,
    numbers=left,
    numbersep=5pt,
    showspaces=false,
    showstringspaces=false,
    showtabs=false,
    tabsize=4,
    frame=single,                 
}

\definecolor{jsonKey}{rgb}{0.08, 0.45, 0.74}    
\definecolor{jsonVal}{rgb}{0.64, 0.08, 0.08}    
\definecolor{jsonNum}{rgb}{0.05, 0.60, 0.45}    
\definecolor{jsonBool}{rgb}{0.00, 0.00, 1.00}   

\lstdefinelanguage{json}{
    keywords={true,false,null},
    keywordstyle=\color{jsonBool}\bfseries,
    sensitive=true,
    string=[b]", 
    stringstyle=\color{jsonVal},
    moredelim=[s][\color{jsonKey}]{"}{:},
    literate=
     *{0}{{{\color{jsonNum}0}}}{1}
      {1}{{{\color{jsonNum}1}}}{1}
      {2}{{{\color{jsonNum}2}}}{1}
      {3}{{{\color{jsonNum}3}}}{1}
      {4}{{{\color{jsonNum}4}}}{1}
      {5}{{{\color{jsonNum}5}}}{1}
      {6}{{{\color{jsonNum}6}}}{1}
      {7}{{{\color{jsonNum}7}}}{1}
      {8}{{{\color{jsonNum}8}}}{1}
      {9}{{{\color{jsonNum}9}}}{1}
      {:}{{:}}{1}      
      {,}{,}{1}        
      {\{}{{{\color{black}\{}}}{1} 
      {\}}{{{\color{black}\}}}}{1}
      {[}{{{\color{black}[}}}{1}
      {]}{{{\color{black}]}}}{1},
}

\lstdefinestyle{json}{
    language=json,               
    basicstyle=\ttfamily\small,  
    breaklines=true,             
    showstringspaces=false,      
    frame=single,                 
    rulecolor=\color{black!30},   
    framesep=10pt,                
    frameround=tttt,              
    numbers=none,                
    keepspaces=true,             
    columns=fullflexible,        
    upquote=true,                
}

\lstdefinestyle{promptstyle}{
    basicstyle=\ttfamily\small, 
    breaklines=true,            
    breakindent=0pt,
    showstringspaces=false,     
    frame=none,                 
    numbers=none,               
    keepspaces=true,            
    columns=fullflexible,       
    tabsize=2,                  
    escapeinside={(*}{*)},  
    moredelim=**[is][\textcolor{blue}]{@}{@},
}

\tcbset {
  base/.style={
    arc=0mm, 
    bottomtitle=-0.25mm,
    boxrule=0mm,
    colbacktitle=black!10!white, 
    coltitle=black, 
    fonttitle=\bfseries, 
    left=2.5mm,
    leftrule=1mm,
    right=3.5mm,
    title={#1},
    toptitle=0.25mm,
    breakable,
  }
}

\tcbset {
  base/.style={
    arc=0mm, 
    bottomtitle=-0.25mm,
    boxrule=0mm,
    colbacktitle=black!10!white, 
    coltitle=black, 
    fonttitle=\bfseries, 
    left=2.5mm,
    leftrule=1mm,
    right=3.5mm,
    title={#1},
    toptitle=0.25mm,
    breakable,
  }
}

\lstdefinelanguage{jsonMemory}{
    keywords={true,false,null},
    keywordstyle=\color{black}\bfseries, 
    sensitive=true,
    string=[b]",
    stringstyle=\color{black},           
    moredelim=[s][\color{black}]{"}{:},
    literate=
      {:}{{:}}{1}
      {,}{,}{1}
      {\{}{{{\color{black}\{}}}{1}
      {\}}{{{\color{black}\}}}}{1}
      {[}{{{\color{black}[}}}{1}
      {]}{{{\color{black}]}}}{1},
}

\lstdefinestyle{memory}{
    language=jsonMemory,         
    basicstyle=\ttfamily\small\color{black}, 
    breaklines=true,
    showstringspaces=false,
    frame=single,
    rulecolor=\color{black!30},
    framesep=10pt,
    frameround=tttt,
    numbers=none,
    keepspaces=true,
    columns=fullflexible,
    upquote=true,
    escapeinside={(*}{*)},
}

\definecolor{brandblue}{rgb}{0.34, 0.7, 1}
\newtcolorbox{mybox}[1]{
  colframe=brandblue,
  base={#1 \hfill \hyperlink{appendixtoc}{\footnotesize\fbox{Back to ToC}}}
}

\definecolor{downred}{HTML}{C0392B} 
\definecolor{upgreen}{HTML}{27AE60}
\definecolor{graytext}{gray}{0.6}

\newcommand{\down}[1]{\textcolor{downred}{\fontsize{6pt}{6pt}\selectfont ($\downarrow$ #1)}}

\newcommand*{\imgintext}[1]{%
  \raisebox{-0.2ex}{%
    \includegraphics[
      height=2.3ex, 
      keepaspectratio
    ]{#1}%
  }%
}

\NewDocumentCommand{\hongru}{ mO{} }{}

\title{Mem$^\textbf{2}$Evolve: Towards Self-Evolving Agents via Co-Evolutionary Capability Expansion and Experience Distillation}

\author{%
Zihao Cheng\textsuperscript{1},
Zeming Liu\textsuperscript{1\dag},
Yingyu Shan\textsuperscript{2},
Xinyi Wang\textsuperscript{3},
Xiangrong Zhu\textsuperscript{3},\\
\textbf{Yunpu Ma\textsuperscript{4},} 
\textbf{Hongru Wang\textsuperscript{5},}
\textbf{Yuhang Guo\textsuperscript{2},} 
\textbf{Wei Lin\textsuperscript{3},}
\textbf{Yunhong Wang\textsuperscript{1},} \\
\textsuperscript{1}Beihang University \quad
\textsuperscript{2}Beijing Institute of Technology \quad
\textsuperscript{3}Independent Researcher \\
\textsuperscript{4}Munich Center for Machine Learning \quad
\textsuperscript{5}University of Edinburgh \\
\textsuperscript{\dag}Corresponding author \quad
Email: \texttt{zihaocheng@buaa.edu.cn}
}

\begin{document}
\startcontents[global]
\maketitle

\begin{abstract}
While large language model--powered agents can self-evolve by accumulating experience or by dynamically creating new assets (i.e., tools or expert agents), existing frameworks typically treat these two evolutionary processes in isolation. This separation overlooks their intrinsic interdependence: the former is inherently bounded by a manually predefined static toolset, while the latter generates new assets from scratch without experiential guidance, leading to limited capability growth and unstable evolution. To address this limitation, we introduce a novel paradigm of co-evolutionary Capability Expansion and Experience Distillation. Guided by this paradigm, we propose the \textbf{Mem$^{\textbf{2}}$Evolve}, which integrates two core components: \textbf{Experience Memory} and \textbf{Asset Memory}. Specifically, Mem$^{2}$Evolve leverages accumulated experience to guide the dynamic creation of assets, thereby expanding the agent's capability space while simultaneously acquiring new experience to achieve co-evolution. Extensive experiments across 6 task categories and 8 benchmarks demonstrate that Mem$^{2}$Evolve achieves improvement of 18.53\% over standard LLMs, 11.80\% over agents evolving solely through experience, and 6.46\% over those evolving solely through asset creation, establishing it as a substantially more effective and stable self-evolving agent framework. Code is available at: \url{https://buaa-irip-llm.github.io/Mem2Evolve}.
\end{abstract}
\section{Introduction}
Large language model (LLM)--powered agents have achieved remarkable success in a wide range of applications~\citep{swe-agent, searchr1, retail, repodebug, toolspectrum}. Building on these successes, recent research is moving beyond static, task-specific systems toward self-evolving agents that can leverage past experiences and autonomously expand their capabilities~\citep{gao2025survey, fang2025comprehensivesurveyselfevolvingai, toa}.

\begin{figure}[t]
    \centering
    \includegraphics[width=\linewidth]{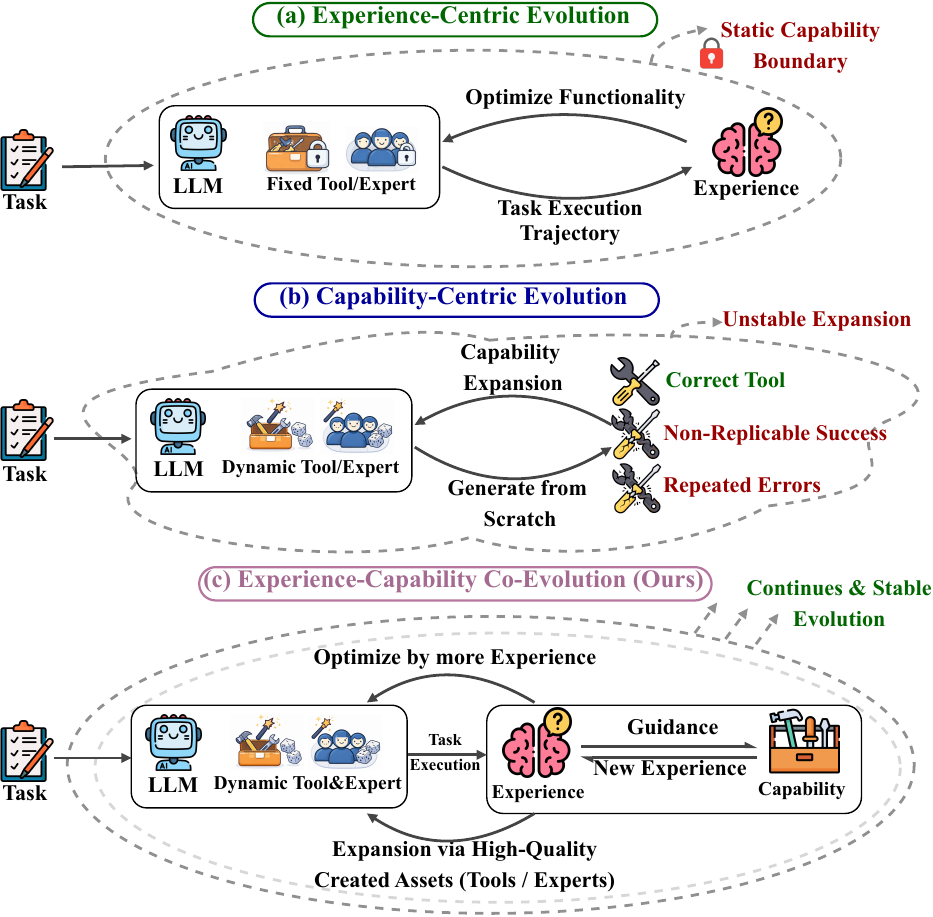}
    \caption{\textbf{Paradigms of Self-Evolving Agents}: (a) Experience-centric evolution, (b) Capability-centric evolution, and (c) Our co-evolutionary framework that jointly expands capabilities and distills experience.}
        
    \label{fig:intro}
\end{figure}
\begin{table*}[t]
\centering
\renewcommand{\arraystretch}{0.79} 

\resizebox{\textwidth}{!}{
\begin{tabular}{@{}l ccc cccc c@{}}
\toprule[0.08em]
\multirow{2.5}{*}{\textbf{Framework}} & 
\multicolumn{3}{c}{\textbf{Experience Distillation}} & 
\multicolumn{4}{c}{\textbf{Capability Expansion}} &
\multirow{2.5}{*}{\textbf{\makecell{Exp.-Guided\\Creation}}} \\ 
\cmidrule(lr){2-4} \cmidrule(lr){5-8}
\renewcommand{\arraystretch}{0.7} 
 & \textbf{Optimization} & \textbf{Persistence} & \textbf{Source} & 
 \textbf{Tool Crea.} & \textbf{Agent Crea.} & \textbf{Tool/Agent} & \textbf{Crea. Grounding} & \\ 
\midrule[0.05em]

DSPy \cite{dspy} & \yes & \no & \imgintext{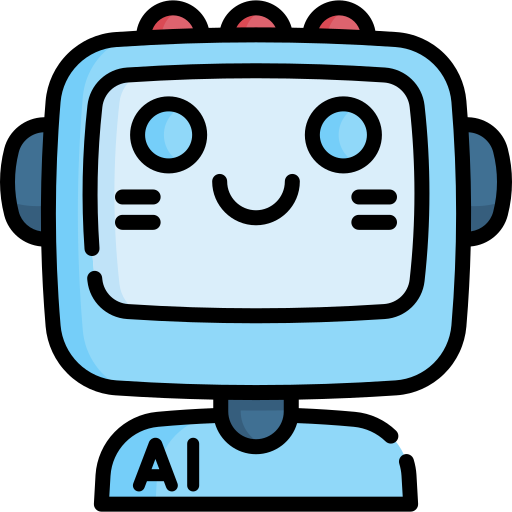} & \no & \no & Static & -- & \no \\
DyLAN \cite{dylan} & \yes & \no & \imgintext{figure/agent.png} & \no & \no & Static & -- & \no \\
ReasoningBank \cite{reasoningbank} & \no & \yes & \imgintext{figure/agent.png} & \no & \no & Static & -- & \no \\
AFlow \cite{aflow} & \yes & \no & \imgintext{figure/agent.png} & \no & \no & Static & -- & \no \\
AgentSquare \cite{agentsquare} & \yes & \no & \imgintext{figure/agent.png} & \no & \no & Static & -- & \no \\
Agentic Neural Networks \cite{ann} & \yes & \no & \imgintext{figure/agent.png} & \no & \no & Static& -- & \no \\
\midrule[0.05em]

AgentVerse \cite{agentverse} & \yes & \no & -- & \no & \yes & Dynamic & \imgintext{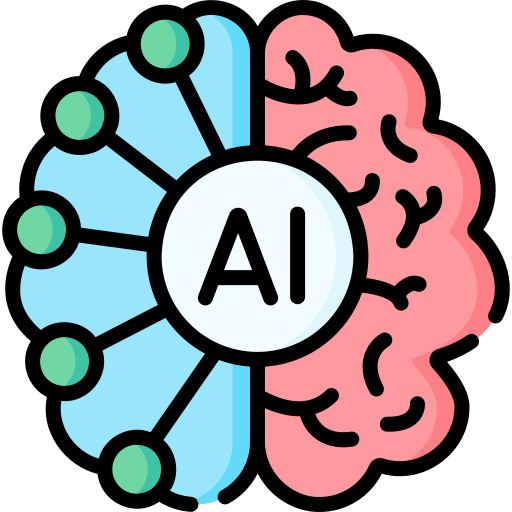} & \no \\
AutoAgents \cite{autoagents} & \no & \no & -- & \no & \yes & Dynamic & \imgintext{figure/parameter.png} & \no \\
SwarmAgentic \cite{swarmagentic} & \yes & \no & -- & \no & \yes & Dynamic & \imgintext{figure/parameter.png} & \no \\
Alita \cite{alita} & \no & \no & -- & \yes & \no & Dynamic& \imgintext{figure/parameter.png}~\texttt{+}~\imgintext{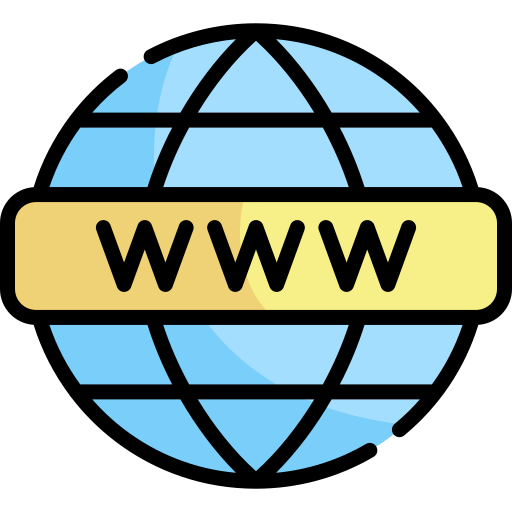} & \no \\
ToolMaker \cite{toolmaker} & \no & \no & -- & \yes & \no & Dynamic& \imgintext{figure/parameter.png}~\texttt{+}~\imgintext{figure/websearch.png} & \no \\

\midrule[0.05em]
\rowcolor{blue!7}
\textbf{Mem$^\textbf{2}$Evolve (Ours)} & \yes & \yes & \imgintext{figure/agent.png}~\texttt{+}~\imgintext{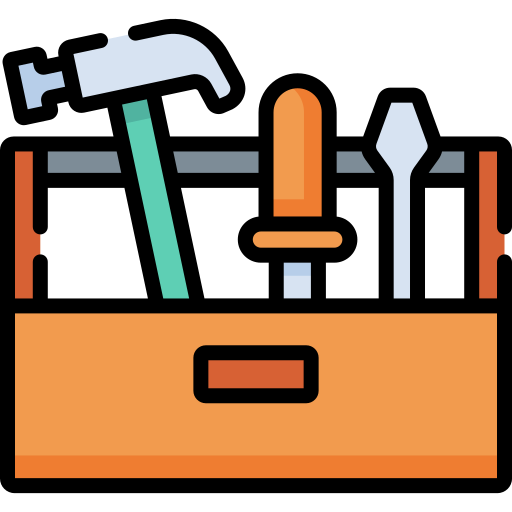} & \yes & \yes & \textbf{Dynamic} & \imgintext{figure/parameter.png}~\texttt{+}~\imgintext{figure/websearch.png}~\texttt{+}~\imgintext{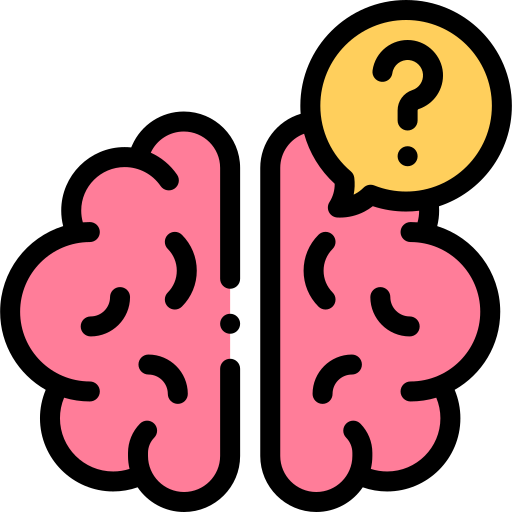} & \yes \\
\bottomrule[0.08em]
\end{tabular}%
}
\caption{\textbf{Comparison of self-evolving agent frameworks.} 
\textbf{Optimization} indicates whether experience is used to optimize the agent (e.g., prompts).
\textbf{Persistence} denotes whether experiences are persistently stored for future reuse.
\textbf{Source}: \imgintext{figure/agent.png} agent task execution trajectory, \imgintext{figure/tool.png} tool creation process.
\textbf{Tool Crea.} and \textbf{Agent Crea.} indicate whether the framework supports creation of tools and expert agents, respectively.
\textbf{Tool/Agent} denotes whether the toolset and expert agents are static or dynamic.
\textbf{Crea. Grounding} indicates the knowledge sources used for asset creation, \imgintext{figure/parameter.png} parametric knowledge, \imgintext{figure/websearch.png} web search information, \imgintext{figure/memory.png} experience.
\textbf{Exp.-Guided Creation} indicates whether new assets are created under the guidance of past experience.
Details in the Appendix~\ref{appendix: defination} and~\ref{appendix: comparasion}.}

\label{tab:comparasion}
\end{table*}

However, current frameworks predominantly treat these evolutionary processes in isolation~\citep{cemri2025multiagentllmsystemsfail}. As illustrated in Figure~\ref{fig:intro}(a), \textbf{Experience-centric evolution}~\citep{evoagent, textgrad} enables systems to learn from experience to optimize execution strategies~\citep{ann}, refine prompts~\citep{aflow}, or build experience repositories~\citep{reasoningbank}. However, this paradigm limits the system to a fixed set of tools and expert agents, leading capability space remains static and cannot expand beyond the pre-specified library. In contrast,  \textbf{capability-centric evolution} (Figure~\ref{fig:intro}(b)) enables the system to dynamically create new tools~\citep{toolmaker, alita} or spawn new expert agents~\citep{agentverse, autoagents, swarmagentic}. However, creating new assets from scratch without the guidance of experience prevents the system from utilizing proven strategies and avoiding known pitfalls, leading to non-replicable success and repeated errors. 

To address these limitations, 
inspired by the equilibrium theory~\citep{piaget1972development}, which posits intelligence evolves through the interplay of assimilation (integrating new experiences) and accommodation (adapting internal structures), 
we introduce a novel paradigm of \textit{co-evolutionary capability expansion and experience distillation} (Figure~\ref{fig:intro}c). In this paradigm, expanding agent capabilities enables it to complete a broader range of tasks, thereby yielding more experiences. These experiences are then distilled to guide subsequent capability expansion, realizing co-evolution of capability and experience.

Guided by this paradigm, we propose \textbf{Mem$^\textbf{2}$Evolve}, an agentic framework that coordinates the evolution of capabilities and experiences through a core dual-memory mechanism comprising Asset Memory and Experience Memory. Specifically, Asset Memory serves as a persistent and extensible repository of the agent's capabilities, organizing expert agents and executable tools. Experience Memory accumulates strategic experience distilled from both successful and failed trajectories to guide future asset creation and task execution. Building upon this dual-memory architecture, Mem$^2$Evolve operates through two complementary phases. During \emph{forward inference}, the system follows a ``reuse first, create on demand'' strategy, leveraging both memories to execute tasks. When a task exceeds the agent's current capability boundary, the system dynamically creates new assets guided by experience to expand its capabilities. Upon task completion, \emph{backward evolution} retains high-quality newly created assets into Asset Memory and distills transferable lessons into Experience Memory. This forward-backward loop enables the co-evolution of capabilities and experiences.

To validate the effectiveness of Mem$^2$Evolve, we conduct extensive experiments across 6 tasks and 8 benchmarks, covering general assistant~\cite{gaia}, multi-hop question answering~\cite{hotpotqa}, mathematical reasoning, embodied task~\cite{alfworld}, planning~\cite{travelplanner}, and web interaction~\cite{webshop}. Beyond achieving superior overall performance against capability- and experience-centric baselines, Mem$^2$Evolve demonstrates robust adaptability, enabling sustained evolution in \textit{single-task} and effective memory reuse in \textit{cross-task} settings.

Our contributions are summarized as follows:

\begin{itemize}[leftmargin=*]
    \setlength{\itemsep}{0.45mm}
    \item To the best of our knowledge, we are the first to propose the co-evolutionary agent paradigm that couples dynamic capability expansion with experience distillation.

    \item Guided by this paradigm, we introduce Mem$^2$Evolve, a dual-memory framework that coordinates Asset Memory for dynamic capability expansion and Experience Memory for strategic experience distillation. Through a forward inference and backward evolution loop, Mem$^2$Evolve continuously leverages and expands both memories, driving capability--experience co-evolution.
        
    \item Extensive experiments show that Mem$^2$Evolve consistently outperforms both capability-centric and experience-centric baselines. Moreover, it exhibits strong adaptability, supporting sustained self-evolution in single-task settings and effective memory reuse for cross-task generalization.
\end{itemize}

\section{Related Work}
\paragraph{Experience-Centric Evolving.} Recent research on self-evolving agents predominantly focuses on optimizing systems by leveraging experience accumulated from past tasks~\citep{evoagent, camel, ann}. For instance, DyLAN~\citep{dylan} and DSPy~\citep{dspy} dynamically select agent teams from a predefined pool by aligning past experience with current task requirements. Similarly, Aflow~\citep{aflow} and AgentSquare~\citep{agentsquare} modularize agents and employ search algorithms to optimize module compositions. ReasoningBank~\citep{reasoningbank} summarizes successful and failed experiences to enhance performance on new tasks. However, as shown in Table~\ref{tab:comparasion}, these frameworks are confined to a fixed set of tools and agents, resulting in a static capability space. Consequently, they cannot extend their boundaries to handle tasks beyond the predefined asset. In contrast, Mem$^2$Evolve dynamically creates high-quality agents and tools, enabling it to transcend these pre-existing capability limits. 

\paragraph{Capability-Centric Evolving.}
In parallel to experience-centric evolving, capability-centric frameworks focus on expanding the boundaries of agentic systems by dynamically generating tools or agents, thereby reducing dependence on manual design~\citep{LATM, adaptive}. AgentVerse~\citep{agentverse} and AutoAgents~\citep{autoagents} generate expert agents tailored to specific task dynamics, extending the system's execution capabilities. ToolMaker~\citep{toolmaker} and Alita~\citep{alita} dynamically create tools to handle videos, documents, and complex mathematical simulations~\citep{retool, wan2026dawn}. However, creating these new assets from scratch without the guidance of experience prevents these systems from leveraging proven strategies or avoiding known pitfalls. This isolation inevitably leads to non-replicable successes and recurring errors. In contrast, Mem$^2$Evolve couples capability expansion with experience distillation, realizing a co-evolution that past insights guide asset creation and new capabilities yield richer experiences.

\section{Mem$^\textbf{2}$Evolve}

\begin{figure*}[t]
    \centering
    \includegraphics[width=\textwidth]{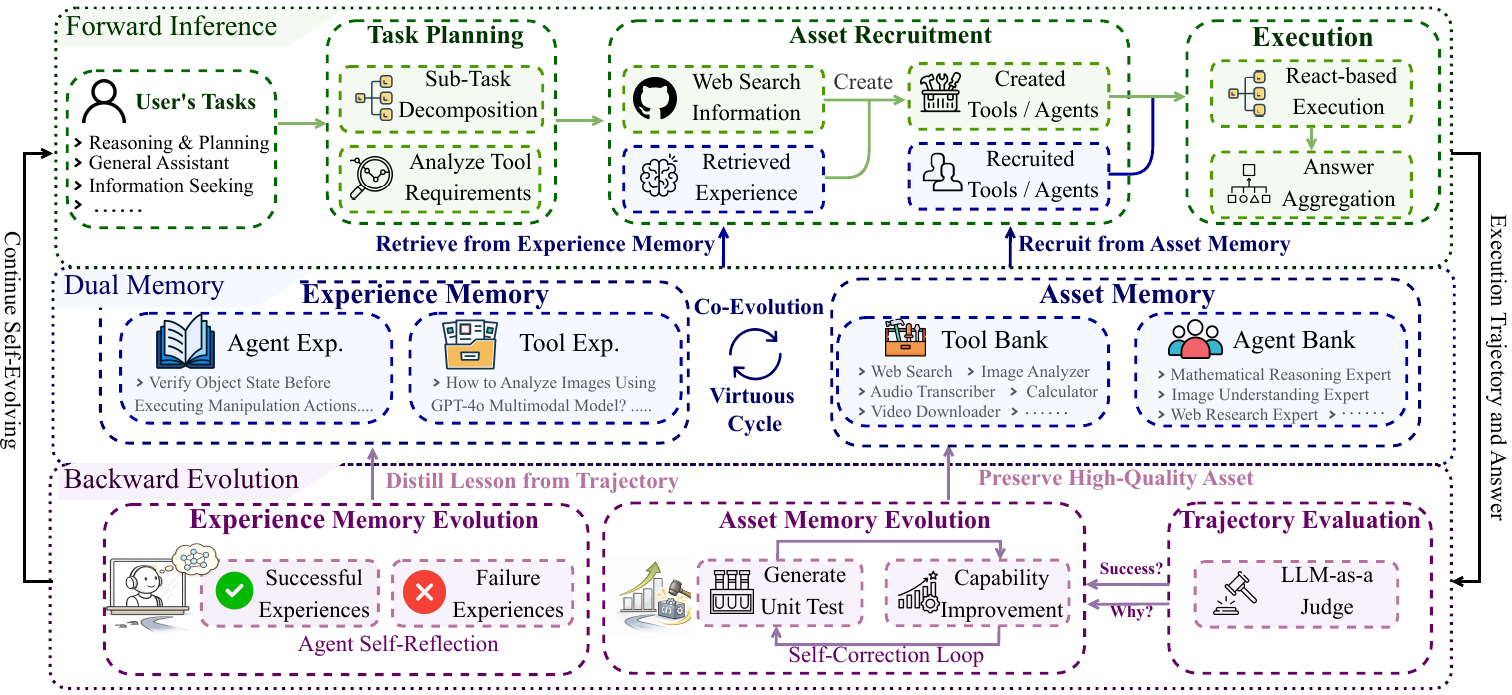}
    \caption{\textbf{Overview of Mem$^\textbf{2}$Evolve}, a self-evolving agent framework built on a \textbf{Dual-Memory} mechanism. The evolution proceeds in two phases. During \textbf{Forward Inference}, the agent recruits tools and expert agents from \textit{Asset Memory} to execute the current task. When the task exceeds its current capability boundary, \textit{Experience Memory} is leveraged to guide the stable creation of new assets on demand. During \textbf{Backward Evolution}, newly validated assets are preserved in Asset Memory to achieve persistent capability expansion, while strategic insights distilled from execution trajectories are accumulated into Experience Memory. This forward–backward loop enables the co-evolution of capabilities and experience, forming a stable self-evolving cycle.}
    \label{fig:method}
\end{figure*}

We present \textbf{Mem$^\textbf{2}$Evolve}, a novel self-evolving agent framework that coordinates capability expansion and experience distillation. As illustrated in Figure~\ref{fig:method}, Mem$^2$Evolve is built upon a \emph{Dual-Memory Mechanism}: \emph{Asset Memory} for dynamic capability expansion and \emph{Experience Memory} for strategic experience distillation(\S\ref{sec:dual-memory}). Built on this dual-memory foundation, Mem$^2$Evolve operates in a two-phase task loop: \emph{forward inference} and \emph{backward evolution}. During \emph{forward inference} (\S\ref{sec:forward}), the agent leverages both memories to execute tasks while dynamically creating new assets to expand its capabilities boundary. Upon task completion, the \emph{backward evolution} process (\S\ref{sec:backward}) retains high-quality assets and distills lessons from execution trajectories, enabling continuous self-evolution.

\subsection{Dual-Memory Mechanism} \label{sec:dual-memory}
We organize the memory into two distinct components: the \textbf{Asset Memory $\mathcal{M}_A$}, which stores the expert agents and tools, and the \textbf{Experience Memory $\mathcal{M}_E$}, which accumulates lessons distilled from past successes and failures to guide future actions.

\subsubsection{Asset Memory}
To support capability expansion at both the strategic level (through expert agents) and the operational level (through tools), Asset Memory serves as a repository of reusable, execution-ready capabilities:
\begin{equation}
    \mathcal{M}_A = \mathcal{B}_{agt} \cup \mathcal{B}_{tool},
\end{equation}
where $\mathcal{B}_{agt}$ is the \emph{Agent Bank} containing expert agents, and $\mathcal{B}_{tool}$ is the \emph{Tool Bank} that stores executable tools.

\paragraph{Agent Bank.}
Building on prior work~\citep{autoagents} and Anthropic's Agent Skills\footnote{\url{https://github.com/anthropics/skills}}, we distill a compact \emph{agent specification} tailored to Mem$^2$Evolve. As exemplified in Figure~\ref{fig:expert agent case}, each entry $m_{agt} \in \mathcal{B}_{agt}$ is defined as:
\begin{equation}
m_{agt} = \langle \rho, \epsilon, \sigma, \mathbb{T}_{avail} \rangle,
\end{equation}
where $\rho$ is the \emph{role} specifying the agent's identity, $\epsilon$ describes the agent's \emph{expertise} and domain knowledge, $\sigma$ denotes \emph{suggestions} that guide the agent's behavior strategies, and $\mathbb{T}_{avail} \subseteq \mathcal{B}_{tool}$ specifies the set of available tools.

\paragraph{Tool Bank.}
To ensure seamless integration with diverse LLM backbones, Tool Bank maintains executable tools stored in compliance with the Model Context Protocol (MCP)\footnote{\url{https://www.anthropic.com/news/model-context-protocol}}, example in Code~\ref{code:piston_sim}. Each entry $m_{tool} \in \mathcal{B}_{tool}$ is defined as:
\begin{equation}
    m_{tool} = \langle n, d_{func}, c_{impl}, \omega_{doc} \rangle,
\end{equation}
where $n$ is the \emph{tool name}, $d_{func}$ provides a \emph{functional description}, $c_{impl}$ contains the \emph{implementation code}, and $\omega_{doc}$ specifies input/output documentation.

\subsubsection{Experience Memory}
To enable Mem$^2$Evolve to replicate proven strategies and circumvent previously encountered pitfalls, Experience Memory accumulates insights derived from past successes and failures, guiding future task execution and asset creation~\cite{reasoningbank}. We define $\mathcal{M}_E = \mathcal{E}_{agt} \cup \mathcal{E}_{tool}$, with each \textit{Memory Item} $e \in \mathcal{M}_E$ structured as: 

\begin{equation}
    e = \langle h_{title}, d_{desc}, \mathcal{U}_{case}, \kappa_{content} \rangle,
\end{equation}
where $h_{title}$ is the title, $d_{desc}$ describes the context, $\mathcal{U}_{case}$ lists applicable use cases, and $\kappa_{content}$ stores the core distilled knowledge, encompassing both agent experience and tool experience:
\paragraph{Agent Experience.}
$\kappa_{content}$ contains strategic insights derived from trajectory reflections, guiding specific expert agents in handling complex tasks.

\paragraph{Tool Experience.}
$\kappa_{content}$ contains implementation guidelines distilled from the tool creation and debugging process, and an example in Figure~\ref{fig: case_tool_experience}.

\subsection{Forward Inference} \label{sec:forward}
To balance the utilization of accumulated expertise with the acquisition of new capabilities, the forward inference follows a strategy of \textit{"Reuse first, Create on demand"}. We formalize it into three phases: \textit{(1) task planning}, \textit{(2) asset recruitment}, and \textit{(3) execution}.

\subsubsection{Task Planning}
Initially, the LLM $\pi_\theta$ acts as a planner to decompose the task $q_t$ into a sequence of sub-tasks $\mathcal{S}=\{s_1,s_2,\dots,s_k\}$, with the prompt in Appendix~\ref{prompt: task planning}. This decomposition ensures that complex problems are broken down into solvable units with clear resource definitions. 

\subsubsection{Asset Recruitment}
For each sub-task $s_i$, the system prepares the required assets via the Recruitment Function $\Gamma(s_i)$:
\begin{small}
\begin{equation}
    \Gamma(s_i) \!=\! 
    \begin{dcases} 
    m^* & \text{sim}(s_i, \mathcal{M}_A) \ge \delta \\
    \text{Create}(s_i \mid \mathcal{M}_E, \text{Web}) & \text{otherwise}
    \end{dcases}
\end{equation}
\end{small}
where $\text{sim}(s_i, \mathcal{M}_A)$ measures the similarity between the sub-task and the asset stored in Assets Memory, 
\hongru{there is no description for agent Bank in asset memory?}
and $\delta$ is a confidence threshold. This mechanism determines whether the sub-task lies beyond the agent's current capability boundary. Depending on the output of $\Gamma(s_i)$, the process branches into two paths:

\paragraph{Recruitment.}
If a high-similarity match exists, the system directly reuses $m^* \in \mathcal{M}_A$. For agents, we select the top-1 candidate surpassing $\delta$ to entrust the sub-task to the most specialized expert. Conversely, for tools, we retrieve the top-$k$ matches to ensure comprehensive utility while mitigating context overhead from excessive documentation.
\hongru{use all assert whose similarity exceeds the threshold? any pre-defined hyper-parameter? e.g., top-k?}

\paragraph{Creation.}
Conversely, for missing capabilities, \textit{Tool Creation} employs experience-augmented generation, conditioning on the sub-task $s_i$, web search results, and relevant experiences $e$ from $\mathcal{E}_{tool}$:
\begin{small}
\begin{equation}
     m_{tool}^{new} \sim \pi_\theta(s_i \mid \text{Retrieve}(s_i, \mathcal{E}_{tool}), \text{Web}(s_i)).
\end{equation}
\end{small}
Similarly, \textit{Agent Creation} synthesizes a new expert by prompting $\pi_\theta$ with task requirements derived from $s_i$, and details in Appendix~\ref{appendix:tool_creation}. 
\hongru{it is a little weird, since if you create agents, you also need to create tool for new agent?}

\subsubsection{Execution}
Each sub-task $s_i$ is assigned to its recruited agent $m_{agt}^i$, augmented with experiences $e$ retrieved from $\mathcal{E}_{agt}$ for role-specific guidance. The agent then executes using available tools $\mathbb{T}_{avail}^i$ within the ReAct framework~\cite{react}, alternating among think, action, and observation steps. Finally, system aggregates all results $\{r_1, \dots, r_k\}$ to produce the final answer $a_t$.

\subsection{Backward Evolution} \label{sec:backward}
Upon task completion, the backward evolution aims to preserve high-quality assets for future reuse and distill transferable lessons from execution trajectories. We formalize it into: \textit{(1) trajectory evaluation, (2) asset memory evolution, and (3) experience memory evolution.}

\subsubsection{Trajectory Evaluation}
The evaluation stage provides the foundation for all subsequent memory updates. We employ an LLM-as-a-Judge~\cite{llmasajudge, reasoningbank, tcmeval} to assess execution quality.\footnote{We assume ground-truth labels are inaccessible during backward evolution to simulate real world. When available, such supervision can further enhance evolution effectiveness.} Given the task $q_t$, execution trajectory $\tau_t$, and answer $a_t$, the Judge produces:
\begin{equation}
    r_t, c_t = \text{Judge}(q_t, \tau_t, a_t),
\end{equation}
where $r_t \in \{0, 1\}$ indicates success or failure, and $c_t$ provides critique comments identifying specific strengths and weaknesses.

\subsubsection{Asset Memory Evolution}
This phase determines which newly created assets should be preserved and refined before entering $\mathcal{M}_A$. Since a correct answer does not guarantee robust underlying assets, we adopt a \emph{Self-Correction Loop} guided by $r_t$ and $c_t$.

For each asset $m_\text{new} \in \mathcal{A}_t^{new}$, where $\mathcal{A}_t^{new}$ denotes the set of newly created assets, we derive a finalized version $m_\text{final}$ as:
\begin{small}
\begin{equation}
    m_\text{final} = 
    \begin{cases} 
    m_\text{new} & \text{if } r_t = 1 \land \\
                 & \text{Valid}(m_\text{new}, c_t) \\[4pt]
    \text{Improve}(m_\text{new}, c_t) & \text{otherwise}
    \end{cases}
    \label{eq:valid function}
\end{equation}
\end{small}
Where $\text{Valid}(m_\text{new}, c_t)$ verifies asset reliability by having the LLM synthesize test cases from the critique $c_t$ and executing $m_\text{new}$ against them. An asset passes validation only if it clears all tests.

If validation fails, $\text{Improve}(m_\text{new}, c_t)$ triggers a \textit{Self-Correction Loop}: revise the asset based on $c_t$ and test failures, then regenerate tests until validation passes. Once validated:
\begin{equation}
    \mathcal{M}_A \leftarrow \mathcal{M}_A \cup \{m_\text{final}\}.
\end{equation}

\begin{table*}[!htbp]
\centering
\setlength{\tabcolsep}{3pt}
\renewcommand{\arraystretch}{1.2}
\resizebox{\textwidth}{!}{%
\begin{tabular}{l *{12}{c}}
\toprule[0.08em]
& \multicolumn{4}{c}{\textbf{GAIA}}
& \multicolumn{1}{c}{\textbf{Embodied}}
& \multicolumn{2}{c}{\textbf{Multi-Hop QA}}
& \multicolumn{2}{c}{\textbf{Math}}
& \multicolumn{1}{c}{\textbf{Planning}}
& \multicolumn{1}{c}{\textbf{Web Interaction}}
& \\

\cmidrule(lr){2-5}
\cmidrule(lr){6-6}
\cmidrule(lr){7-8}
\cmidrule(lr){9-10}
\cmidrule(lr){11-11}
\cmidrule(lr){12-12}

\multirow{-2}{*}{\textbf{Method}}
& \textbf{L1} & \textbf{L2} & \textbf{L3} & \textbf{Total}
& \textbf{ALFWorld}
& \textbf{HotpotQA} & \textbf{2Wiki}
& \textbf{AIME24} & \textbf{AIME25}
& \textbf{TravelPlanner}
& \textbf{WebShop}  
& \multirow{-2}{*}{\textbf{Avg.}} \\

\midrule[0.05em]
\rowcolor{blue!7}	
\multicolumn{13}{c}{\textit{\textbf{Naive-Large Language Model}}} \\
\midrule[0.05em]
GPT-5-Chat (Direct) & 16.98 & 12.79 & 7.69 & 12.49 & 83.58 & 50.40 & \underline{81.80} & 60.00 & 46.67 & 38.68 & 22.31 & 49.49 \\
GPT-5-Chat (CoT)    & 24.53 & 17.44 & 11.54 & 17.84 & 83.58 & 47.40 & 74.40 & 66.67 & 56.67 & 39.51 & 27.49 & 51.71 \\
GPT-5-Chat (ReAct)  & 26.42 & 17.44 & 11.54 & 18.47 & 86.87 & 41.40 & 48.40 & 66.67 & 60.00 & 39.13 & 25.10 & 48.27 \\
OpenAI-DeepResearch$^{\dagger}$ & 74.29 & 69.06 & \underline{47.60} & 67.36 & --- & --- & --- & --- & --- & --- & --- & --- \\
\midrule[0.05em]
\rowcolor{blue!7}	
\multicolumn{13}{c}{\textit{\textbf{Experience-Centric Evolving}}} \\
\midrule[0.05em]

DyLAN    & 24.53 & 19.78 & 11.54 & 18.62 & 91.20 & 52.00 & 65.00 & 46.67 & 43.33 & 43.15 & 36.40 & 49.55 \\
EvoAgent & 22.64 & 19.78 & 11.54 & 17.99 & 92.50 & 54.40 & 75.00 & 66.67 & 43.33 & 49.20 & 37.80 & 54.61 \\
AFLOW    & 26.42 & 17.44 & 15.38 & 19.75 & \underline{93.40} & \textbf{60.80} & 72.40 & 66.67 & 63.33 & 53.24 & \underline{37.90} & 58.44 \\
DSPy     & 30.19 & 15.12 & 11.54 & 18.95 & 92.80 & 55.60 & 76.40 & 66.67 & 50.00 & 44.90 & 35.50 & 55.10 \\
\midrule[0.05em]
\rowcolor{blue!7}	
\multicolumn{13}{c}{\textit{\textbf{Capability-Centric Evolving}}} \\
\midrule[0.05em]

Alita        & \underline{81.13} & \underline{75.58} & 46.15 & \underline{72.73} & 86.13 & \underline{58.80} & 77.40 & \underline{70.00} & \underline{66.67} & 48.32 & 30.21 & \underline{63.78} \\
AgentVerse   & 30.19 & 16.28 & 19.23 & 21.90 & 88.32 & 38.60 & 74.60 & 60.00 & 50.00 & 47.25 & 32.53 & 51.65 \\
AutoAgens    & 35.85 & 24.42 & 19.23 & 26.50 & 87.92 & 54.20 & 73.80 & 40.00 & 36.67 & 43.52 & 31.40 & 49.25 \\
SwarmAgentic & 28.30 & 18.60 & 13.46 & 20.40 & 88.79 & 56.00 & 80.00 & 46.67 & 40.00 & \underline{59.14} & 34.12 & 53.14 \\

\midrule[0.05em]
\rowcolor{blue!7}	
\multicolumn{13}{c}{\textit{\textbf{Ours}}} \\
\midrule[0.05em]

\textbf{Mem$^\textbf{2}$Evolve} & \textbf{88.68} & \textbf{82.56} & \textbf{57.69} & \textbf{76.31} & \textbf{94.31} & \textbf{60.80} & \textbf{82.00} & \textbf{76.70} & \textbf{73.33} & \textbf{59.25} & \textbf{39.20} & \textbf{70.24} \\
\bottomrule[0.08em]
\end{tabular}
}
\caption{\textbf{Main results across 6 tasks and 8 benchmarks}, reported as Pass@1 for each benchmark. The best results are highlighted in \textbf{bold}, and the second-best results are \underline{underlined}. $^{\dagger}$Results are from the original paper.}
\label{tab:main_results}
\end{table*}

\subsubsection{Experience Memory Evolution}
Mem$^2$Evolve distills trajectory-level insights into the $\mathcal{M}_E$ to guide future task execution and asset creation. After each task, the system reflects on the trajectory $\tau_t$ and $(r_t, c_t)$ to extract \textit{Memory Items}:
\begin{equation}
    e_\text{new} = \text{Reflection}(\tau_t, r_t, c_t),
\end{equation}
The Reflection function captures insights from both successful and failed executions.

\paragraph{Success Generalization.}
When $r_t = 1$, Mem$^2$Evolve abstracts high-level guidelines from the successful trajectory. For agents, $\kappa_{content}$ records strategic advice and coordination patterns for specific roles; for tools, it captures effective implementation patterns and usage recipes.

\paragraph{Failure Diagnosis.}
When $r_t = 0$ or $c_t$ indicates substantial debugging, Reflection focuses on failure modes. The resulting $e_\text{new}$ encodes anti-patterns and failure–fix pairs to prevent similar errors. Detailed prompt in Appendix~\ref{prompt:tool memory}~and~\ref{prompt:agent_memory}.

Finally, the distilled experience items are merged into the Experience Memory:
\begin{equation}
    \mathcal{M}_E \leftarrow \mathcal{M}_E \cup \{e_\text{new}\}.
\end{equation}

\section{Experiments}

\subsection{Experiment Setting}
\paragraph{Baselines.}
Following prior research~\cite{alita, swarmagentic}, we compare Mem$^2$Evolve against three categories of baselines: (1) \textbf{Naive LLMs}, including Direct prompting, CoT~\cite{cot}, ReAct~\cite{react}, and OpenAI's DeepResearch~\citep{deepreasearch}, (2) \textbf{Experience-Centric frameworks} such as DyLAN~\cite{dylan}, EvoAgent~\cite{evoagent}, AFLOW~\cite{aflow}, and DSPy~\cite{dspy}, and (3) \textbf{Capacity-Centric frameworks}, spanning tool-generative methods Alita~\cite{alita}, ToolMaker~\cite{toolmaker}) and agent-generative approaches AgentVerse~\cite{agentverse}, AutoAgents~\cite{autoagents}, SwarmAgentic~\cite{swarmagentic}. More details in the Appendix~\ref{appendix:baselines}.

\paragraph{Benchmarks.}
Following \citet{deepagent, otcpo}, we evaluate the agent's capabilities across 8 benchmarks in 6 distinct tasks. These include \textbf{GAIA}~\cite{gaia} for general assistant, \textbf{ALFWorld}~\cite{alfworld} and \textbf{WebShop}~\cite{webshop} for embodied and web interaction, and \textbf{TravelPlanner}~\cite{travelplanner} for planning. We also include \textbf{HotpotQA}~\cite{hotpotqa} and \textbf{2WikiMultihopQA}~\cite{2wiki} for multi-hop QA, plus \textbf{AIME 24/25} for mathematical reasoning. Details are in Appendix~\ref{appendix:benchmarks}.

\paragraph{Implement Details.}
For all baselines, we utilize GPT-5-chat\footnote{\url{https://openai.com/index/introducing-gpt-5/}} as the LLM backbone. The web search tool incorporates the Serper search engine\footnote{\url{https://serpapi.com/}} and the Crawl4AI~\cite{crawl4ai} parsing framework, and code execution is managed via the SandboxFusion environment~\cite{sandboxfusion}.

\subsection{Main Results}
Table \ref{tab:main_results} presents the comparative results of different frameworks, and the following conclusions are derived based on these results.

\paragraph{\textit{Capability-experience co-evolution achieves the strongest general agent.}}
Mem$^2$Evolve achieves the best overall performance among all evaluated frameworks, demonstrating the effectiveness of jointly evolving capabilities and experience. Under the same GPT-5-chat as all baselines, Mem$^2$Evolve attains an average Pass@1 of 70.24\% across all benchmarks, outperforming the strongest capability-centric baseline Alita by 6.46\%, experience-centric baseline Aflow by 11.80\%, and naive-llm by up to 18.53\%. These consistent improvements confirm that capability–experience co-evolution yields a substantially more powerful general agent than either paradigm.

\paragraph{\textit{Breaking the Capability Boundaries of Static Agents.}}
When initialized with only a Web Search tool, purely experience-centric methods yield marginal improvements over the base LLM. On GAIA, experience-centric baselines with a fixed toolset improve Pass@1 by at most 1.28\%; on AIME, AFLOW achieves only +3.33\% on AIME25 and no improvement on AIME24. In contrast, Mem$^2$Evolve, starting from the same minimal configuration but capable of evolving new tools and expert agents, achieves +57.84\% on GAIA and +10.03\%/+13.33\% on AIME24/AIME25, respectively. These substantial gains demonstrate that Mem$^2$Evolve effectively extends the capability boundary of the base LLM.

\paragraph{\textit{Experience Memory Enhances Capability Expansion.}}
Incorporating Experience Memory further enhances the effectiveness of capability expansion. Under matched conditions, Mem$^2$Evolve outperforms the capability-centric baseline Alita by 6.46\% in average Pass@1. This improvement suggests that Experience Memory refines and stabilizes the utilization of newly evolved tools and agents, enabling capability expansion to translate more reliably into downstream performance gains.

\section{Analysis}
In this section, we conduct a comprehensive analysis to answer the following research questions \textbf{RQ1:} \textit{What role does each module play in Mem$^2$Evolve?} (\S\ref{sec:ablation}) \textbf{RQ2:} \textit{How does experience guide asset generation?} (\S\ref{sec:experience-guide}) \textbf{RQ3:} \textit{How does Mem$^2$Evolve self-evolve in single task?} (\S\ref{sec:single}) \textbf{RQ4:} \textit{How does Mem$^2$Evolve self-evolve across tasks?} (\S\ref{sec:cross})  \textbf{RQ5:} \textit{How does Mem$^2$Evolve behave in case studies?} (\S\ref{appendix: case_rq})

\subsection{RQ1: Ablation Study} \label{sec:ablation}
\begin{table}[!htbp]
\centering
\small
\renewcommand{\arraystretch}{0.95}
\setlength{\tabcolsep}{4pt}

\resizebox{0.95\columnwidth}{!}{
\begin{tabular}{@{}l|c|c@{}}
\toprule[0.08em]
\textbf{Framework} & \textbf{Avg. Pass@1} & \textbf{$\Delta_{\textbf{rel}}^\textbf{\%}$} \\
\midrule[0.05em]
\textbf{Mem$^\textbf{2}$Evolve} & \textbf{70.24} & -- \\
\midrule[0.05em]
\rowcolor{blue!7}	
\multicolumn{3}{l}{\textit{\textbf{w/o Asset Creation}}} \\
\midrule[0.05em]
\hspace{3mm} w/o Tool Creation & 59.96 & $\downarrow 10.28$ \\
\hspace{3mm} w/o Expert Agent Creation & 68.52 & $\downarrow 1.72$ \\

\midrule[0.05em]
\rowcolor{blue!7}	
\multicolumn{3}{l}{\textit{\textbf{w/o Experience Distillation}}} \\
\midrule[0.05em]
\hspace{3mm} w/o Tool Memory & 67.11 & $\downarrow 3.13$ \\
\hspace{3mm} w/o Agent Memory & 65.51 & $\downarrow 4.73$ \\

\bottomrule[0.08em]
\end{tabular}
}
\caption{\textbf{Ablation study of Mem$^\textbf{2}$Evolve.} Full results are provided in Appendix~\ref{tab:ablation_results}.}
\label{tab:ablation_study}
\end{table}
To verify the effectiveness of each module in Mem$^2$Evolve, we conducted an ablation study on Asset Creation and Experience Distillation. As shown in Table~\ref{tab:ablation_study}, Mem$^2$Evolve consistently outperforms all variants, validating the necessity of the proposed Dual-Memory mechanism. Specifically, \textit{w/o Tool Creation} causes the largest performance drop of 10.28\%, highlighting that dynamically expanding the toolset is crucial for handling complex tasks, while \textit{w/o Expert Agent Creation} still leads to a 1.72\% decline because all tasks are forced onto a single general-purpose agent rather than expert agents. Moreover, removing \textit{Agent Memory} causes a 4.73\% performance drop, and removing \textit{Tool Memory} causes a 3.13\% performance drop, as this prevents the system from leveraging validated successes and past failures during both tool creation and task execution, making it difficult to reliably reproduce effective behaviors and avoid known mistakes, thereby degrading overall performance.

\subsection{RQ2: Experience-Guided Asset Creation} 
\label{sec:experience-guide}
\begin{table}[!htbp]
\centering
\small
\renewcommand{\arraystretch}{0.95}
\resizebox{0.95\columnwidth}{!}{%
\begin{tabular}{@{}l|c|c|c@{}}
\toprule[0.08em]
\textbf{Benchmark} & \textbf{w/o Exp.-Guide} & \textbf{w/ Exp.-Guide} & \textbf{$\Delta_{\text{\textbf{rel}}}^{\textbf{\%}}$} \\
\midrule[0.05em]
\rowcolor{blue!7}	
\multicolumn{4}{l}{\textit{\textbf{First-Pass Validity ($\uparrow$)}}} \\
\midrule[0.05em]
\hspace{3mm} GAIA & 32.7\% & 51.0\% (+18.3\%) & $\uparrow$ 56.0 \\
\hspace{3mm} AIME24 & 64.9\% & 83.8\% (+18.9\%) & $\uparrow$ 29.1 \\
\hspace{3mm} AIME25 & 61.8\% & 82.4\% (+20.6\%) & $\uparrow$ 33.3 \\
\hspace{3mm} Avg. & 53.1\% & 72.4\% (+19.3\%) & $\uparrow$ 36.3 \\
\midrule[0.05em]
\rowcolor{blue!7}	
\multicolumn{4}{l}{\textit{\textbf{Avg. Improve Iter. ($\downarrow$)}}} \\
\midrule[0.05em]
\hspace{3mm} GAIA & 1.45 & 0.94 (-0.51) & $\downarrow$ 35.2 \\
\hspace{3mm} AIME24 & 0.76 & 0.24 (-0.52) & $\downarrow$ 68.4 \\
\hspace{3mm} AIME25 & 0.82 & 0.26 (-0.56) & $\downarrow$ 68.3 \\
\hspace{3mm} Avg. & 1.01 & 0.48 (-0.53) & $\downarrow$ 52.5 \\
\bottomrule[0.08em]
\end{tabular}
}
\caption{\textbf{Analysis of Experience-Guided Asset Creation.} We report first-pass validity and average improvement iterations on benchmarks requiring extensive tool generation. Experience guidance is associated with higher first-pass validity, with a relative improvement of up to 56.0\% on GAIA, and fewer fix iterations, with reductions of nearly 68\% on AIME benchmarks.}
\label{tab:visual_analysis}
\end{table}

In Section~\ref{sec:ablation}, we show that incorporating \textit{Tool Memory} leads to consistent performance improvements. This section further investigates its impact on benchmarks that require extensive tool creation. We evaluate experience guidance using: (1) \textit{First-Pass Validity}, which measures whether the initially generated tool satisfies the verification function $\text{Valid}(m_\text{new}, c_t)$ in Equation~\ref{eq:valid function}, and (2) \textit{Avg. Improve Iter.}, defined as the average steps of $\text{Improve}(m_\text{new}, c_t)$ during the self-correction loop. 

As shown in Table~\ref{tab:visual_analysis}, experience guidance substantially improves the reliability and efficiency of tool creation. For AIME24/25, where the agent already demonstrates strong performance, experience guidance reduces the average number of debugging iterations by nearly 68\% and increases first-pass validity to over 82\%, indicating more accurate tool generation at the initial attempt. In contrast, in the more complex GAIA, experience guidance improves first-pass validity by 56.0\% relative to the w/o Exp-Guide, suggesting that experience effectively constrains tool generation toward feasible solutions. Results and case in Figure~\ref{fig:case4} show that experience guidance significantly enhances the stability of tool generation, ensuring a more robust evolutionary trajectory for the agent.

\subsection{RQ3: Single Task Self-Evolving} \label{sec:single}
\begin{figure}[!htbp]
    \centering
    \includegraphics[width=0.9\linewidth]{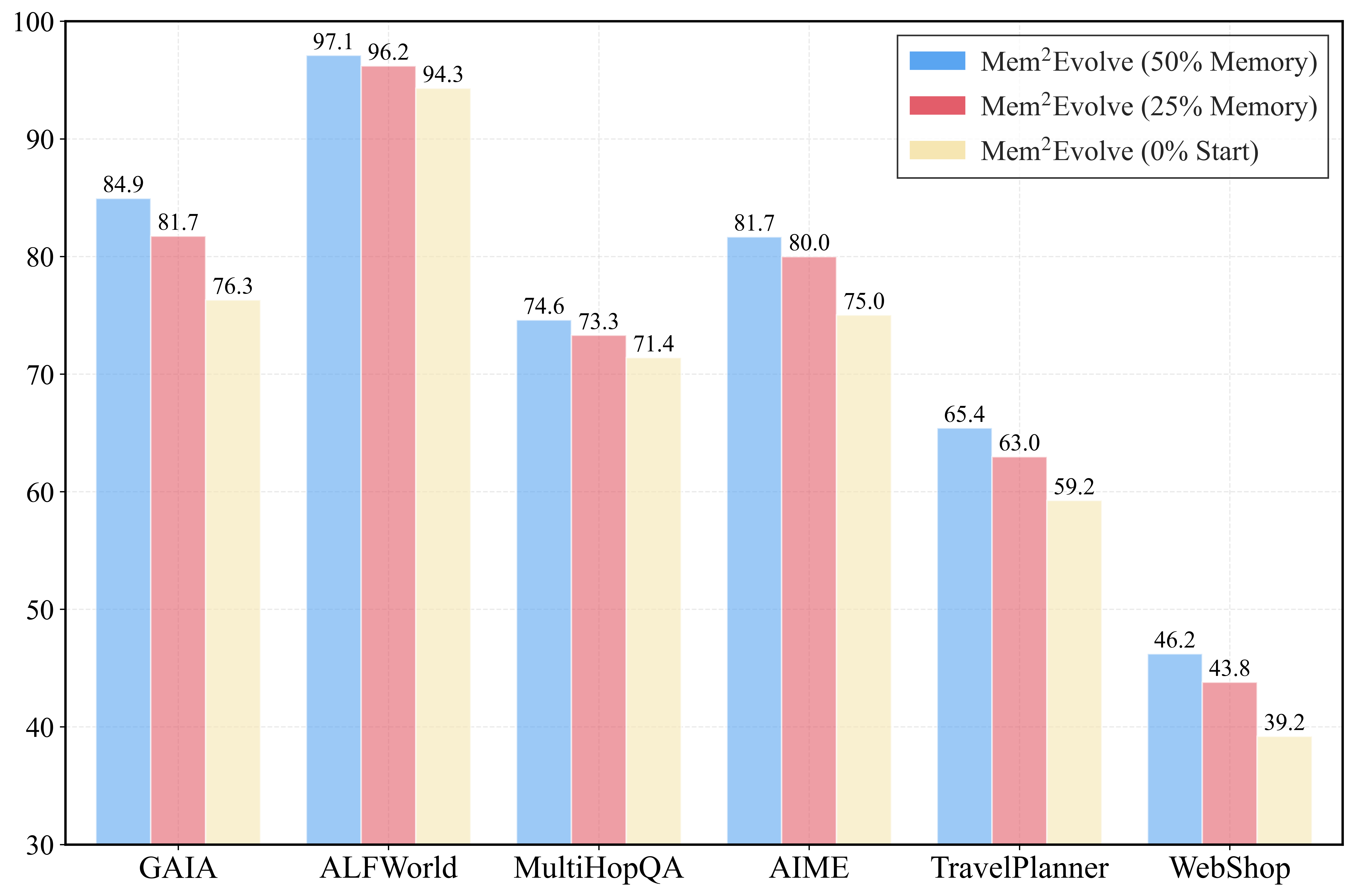}
    \caption{\textbf{Single-task self-evolving performance.} The results show that initializing the agent with prior memory consistently improves performance compared to the setting without initial memory, indicating that Mem$^2$Evolve can effectively leverage accumulated experience to enhance the task execution performance.}
    \label{fig:rq3}
\end{figure}
In Table~\ref{tab:main_results}, we evaluate the performance of Mem$^2$Evolve across multiple benchmarks, where each run starts without any pre-existing memory except for access to the web search tool. In this section, we further analyze the effect of introducing initial memory within the same task. Specifically, we construct initial memory using a subset of data from each benchmark and use it to initialize the system, after which evaluation is conducted on the remaining test set.

As shown in Figure~\ref{fig:rq3}, introducing initial memory consistently improves performance across all benchmarks compared to the setting without initial memory. Most of the performance gains are achieved with a relatively small amount of initial memory, while further enlarging the memory yields diminishing incremental improvements. This pattern suggests that memory accumulated within the same task is effective in enhancing agent performance, with early-stage memory capturing a large fraction of broadly applicable assets and high-utility experience.

\subsection{RQ4: Cross Tasks Self-Evolving} \label{sec:cross}
\begin{figure}[!htbp]
    \centering
    \includegraphics[width=\linewidth]{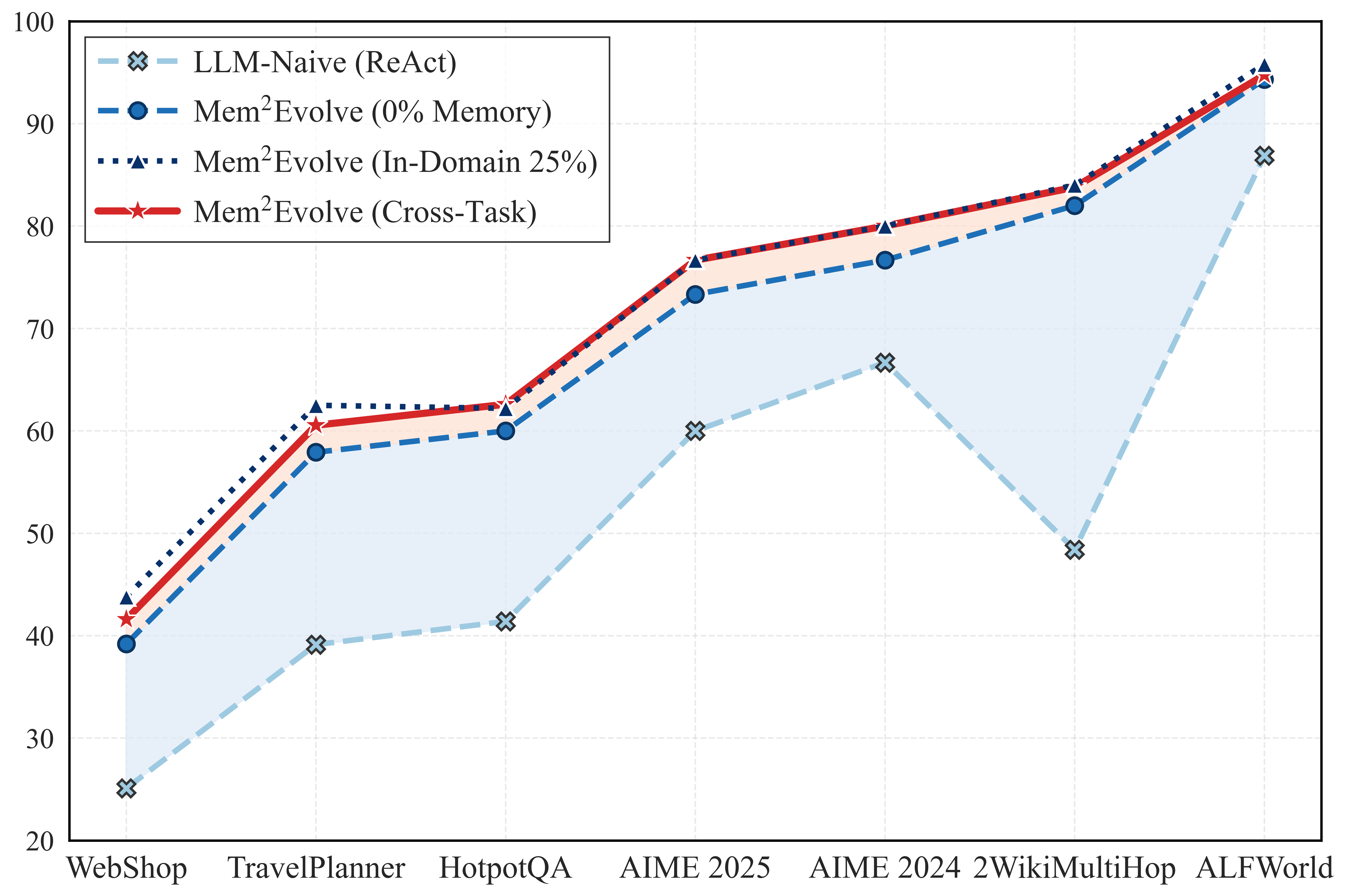}
    \caption{\textbf{Cross-task self-evolving performance.} When initialized with heterogeneous memory from GAIA, Mem$^2$Evolve consistently outperforms the setting without initial memory and achieves performance comparable to single-task initialization.}
    \label{fig:rq4}
\end{figure}
To evaluate the generalization capability of Mem$^2$Evolve in a cross-task setting, we initialize the agent with heterogeneous memory accumulated from GAIA and evaluate its performance across 7 target benchmarks.

As illustrated in Figure~\ref{fig:rq4}, cross-task memory initialization consistently improves performance compared to the setting without initial memory, and achieves results comparable to the 25\% single-task initialization. Despite the domain mismatch between source and target tasks, the agent maintains stable evolutionary trajectories without suffering negative transfer. These results suggest that Mem$^2$Evolve can reuse heterogeneous memory across tasks without adversely affecting performance. The structured representation of memory components and the retrieval mechanism contribute to this behavior by enabling selective access to task-relevant information.

\section{Conclusion}
We introduce Mem$^2$Evolve, a self-evolving agent framework that integrates Asset Memory and Experience Memory and enables their coordinated co-evolution. This design allows the agent to expand its capability space while continuously accumulating strategic experience, leading to more stable and sustained performance improvements. Extensive experimental results show that Mem$^2$Evolve consistently improves performance in both single-task and cross-task settings. We hope that Mem$^2$Evolve provides a practical foundation for building general-purpose, lifelong-learning agents with reduced reliance on human intervention.

\section*{Limitations}
Mem$^2$Evolve is a self-evolving agent framework equipped with both asset memory and experience memory. During task execution, the framework dynamically generates expert agents and tools guided by past experience, thereby continuously expanding its capability boundaries while leveraging past experience to facilitate current task execution and achieve stable self-evolution. However, Mem$^2$Evolve relies on a sandbox environment to execute this autonomously generated code. This dependency limits the system's deployment scope, such as in open-world environments that require direct interaction with local file systems or unrestricted network access.

\bibliography{custom}
\clearpage
\appendix
\onecolumn
\hypertarget{appendixtoc}{}
\printcontents[global]{l}{1}{\setcounter{tocdepth}{2}}
\clearpage

\section{Mem$^\textbf{2}$Evolve} \label{sec:appendix}
\appsection{Defining Continuous and Stable Evolving} \label{appendix: defination}
We define two fundamental characteristics of self-evolving agents: \textbf{(1)} the ability to continuously evolve with minimal human intervention by persistently expanding their capabilities to solve unseen tasks, and \textbf{(2)} the capability to efficiently leverage past experience, enabling correct experience transfer and effective error avoidance for seen tasks, either offline or online.

\begin{itemize}
    \item \textbf{Optimization.} An agent system should be capable of automatically optimizing its internal instructions and coordination workflows based on environmental feedback, thereby achieving optimal task-specific performance. Traditional agent development paradigms rely heavily on manual prompt engineering or predefined interaction protocols. Such approaches are not only labor-intensive but also poorly suited to dynamically evolving task requirements. Therefore, an effective self-evolution framework should emulate a form of “backpropagation” mechanism, continuously refining agent role definitions, prompting strategies, and even multi-agent collaboration topologies through feedback signals or textual gradients derived from task execution outcomes. Crucially, this optimization should not be limited to transient runtime adjustments but should fundamentally enhance the system’s intrinsic competence in handling similar tasks. 

    \item \textbf{Experience Persistence.} To enable genuine lifelong learning, the framework must be able to transform both successful strategies and failure cases from historical tasks into long-term memory assets that persist beyond the lifecycle of a single task. Many existing methods reset system states after task completion, forcing agents to re-explore from scratch when encountering similar scenarios. This not only wastes computational resources but also allows recurring errors. Hence, a cross-task experience persistence mechanism is essential. Whether implemented via explicit external databases that store reasoning trajectories or via implicit knowledge internalization through parameter or prompt updates, this mechanism should enable rapid retrieval and reuse of prior knowledge when facing new tasks, thereby mitigating the cold-start problem and avoiding known pitfalls.

    \item \textbf{Agent Creation.} The framework should not depend on predefined expert agent modules with fixed roles, prompts, or decision logics. Instead, it should be capable of dynamically constructing optimal expert agent teams conditioned on task demands. This capability is critical for high-level, complex planning tasks, where increasing task complexity typically entails decomposing the problem into multiple subtasks with distinct objectives. A single general-purpose agent is often insufficient to handle all subtasks effectively, necessitating specialized agents that each address the components they are best suited for. However, given the vast diversity of real-world tasks, manually predefined expert agents cannot cover all possible scenarios. The system must therefore support dynamic, task-driven agent generation.

    \item \textbf{Tool Creation.} By invoking tools, agent systems can substantially expand their capability boundaries and overcome the limitations imposed by static knowledge. Tools enable access to real-time information, execution of complex mathematical reasoning, and completion of specialized operations. However, task-specific tools typically require careful manual design. When confronted with general, previously unseen tasks, human developers are often still required to create new tools, which is inherently unscalable. To enable continual capability expansion, the framework must therefore possess the ability to autonomously generate tools.

    \item \textbf{Experience-Guided Creation.} When encountering unseen tasks that require the generation of new agents or tools, the framework should leverage its internal assets and accumulated memory to guide the creation process. For example, when a new task involves parsing YouTube video subtitles, previously generated tool documentation for downloading YouTube videos can serve as a reference to facilitate new tool construction. This experience-guided mechanism improves the stability of generated tools and agents, reduces randomness and hallucinations in large language model outputs, and thereby enables a more robust and reliable evolutionary process.
\end{itemize}

\appsection{Evaluation of Existing Self-Evolving Agent Frameworks} \label{appendix: comparasion}

\begin{itemize}
    \item \textbf{DyLAN}~\citep{dylan} models multi-agent collaboration as a Temporal Feed-Forward Network, implementing a "Team Optimization" stage that utilizes a backward message-passing mechanism to calculate "Agent Importance Scores" based on unsupervised peer ratings. \textbf{DyLAN satisfies Optimization:} it actively employs environmental feedback to refine the collaboration topology iteratively. By identifying and selecting the most contributory agents while pruning low-performing ones, DyLAN automatically optimizes the team composition and interaction structure for specific tasks, aligning with the definition of optimizing collaboration logic and topology. However, \textbf{DyLAN fails Agent Creation and Tool Creation:} it does not generate new expert definitions or tools from scratch; instead, it relies on selecting the best subset from a fixed, pre-defined candidate pool of agents. Finally, it \textbf{offers limited Experience Persistence:} while it can reuse calculated importance scores for similar tasks, it lacks a semantic memory mechanism to guide the generation of new assets for entirely unseen domains.

    \item \textbf{DSPy}~\citep{dspy} introduces a programming model that abstracts language model pipelines as text transformation graphs, allowing users to define declarative modules (e.g., ChainOfThought) via natural language signatures. \textbf{DSPy satisfies Optimization}: it employs a compiler with various ``teleprompters'' to automatically refine the pipeline's instructions or fine-tune the underlying language model weights based on metric-driven feedback and bootstrapped demonstrations. However, \textbf{DSPy fails Agent Creation and Tool Creation}: it relies on the user to explicitly define the program structure, the flow of control, and the specific modules and tools to be used, rather than autonomously synthesizing new agent roles or executable tools from scratch. Furthermore, \textbf{DSPy fails Experience Persistence}: it treats experience utilization as a discrete compilation process rather than a continuous memory accumulation. Once the pipeline is compiled, the historical traces are frozen into static few-shot examples or weights, lacking a dynamic, retrievable memory bank to persistently store and reuse new inference trajectories for future tasks.

    \item \textbf{ReasoningBank}~\citep{reasoningbank} introduces a memory framework that distills generalizable reasoning strategies from both successful and failed trajectories, enabling agents to retrieve relevant insights for new tasks. \textbf{ReasoningBank satisfies Experience Persistence}: it explicitly stores abstracted reasoning patterns in a long-term memory bank, allowing the system to mitigate the cold-start problem by transferring knowledge across tasks and preventing the repetition of past errors. However, \textbf{ReasoningBank fails Optimization}: while it improves performance via RAG, it does not structurally optimize the agent's topology, internal prompt templates, or parameters. The agent's core functionality remains static, relying on external memory injection rather than internal refinement. It likewise fails \textbf{Agent Creation} and \textbf{Tool Creation}, as it operates with a fixed agent architecture (e.g., ReAct), leveraging dynamic memory, rather than generating new agent entities or executable tools from scratch. Consequently, it also fails \textbf{Experience-Guided Creation}, as there is no asset generation process to be guided by its rich memory.

    \item \textbf{AFlow}~\citep{aflow} reformulates agentic system development as a search problem over code-represented workflows, utilizing Monte Carlo Tree Search (MCTS) to iteratively explore and optimize the design space. \textbf{AFlow satisfies Optimization}: it treats the workflow structure and node-level prompts as hyperparameters, optimizing them based on execution feedback (e.g., success rates, costs) to find the most effective graph topology. It also satisfies \textbf{Agent Creation} and \textbf{Experience-Guided Creation}: the framework autonomously constructs new workflow nodes (effectively new agents) and connections by leveraging the search history (MCTS values) to guide the generation process, moving away from manual engineering. However, \textbf{AFlow fails Tool Creation}: it focuses on orchestrating the flow of LLM calls and existing tools rather than synthesizing new executable tool code from scratch. Furthermore, it fails \textbf{Experience Persistence}: the experience is utilized only during the offline search phase to produce a static, compiled workflow; it lacks a dynamic, long-term memory mechanism to continuously accumulate and retrieve reasoning patterns for lifelong learning across different tasks.

    \item \textbf{AgentSquare}~\citep{agentsquare} proposes a search-based framework that automates the design of LLM agents by exploring a modular space comprising planning, reasoning, tool use, and memory modules. \textbf{AgentSquare satisfies Optimization}: it treats the agent design process as an objective function maximization problem, iteratively refining the agent's architecture (i.e., module combinations) based on evaluation feedback to find the optimal configuration. It also satisfies \textbf{Agent Creation} and \textbf{Experience-Guided Creation}: the framework autonomously synthesizes new agent instances by recombining modules and utilizes an ``Experience Pool'' (containing history of evaluated agent-task pairs) to train a surrogate model, which efficiently guides the generation of new candidates towards high-performance regions. It further satisfies \textbf{Experience Persistence} by explicitly storing these search trajectories and evaluation results in the Experience Pool, enabling the system to learn from past search iterations. However, \textbf{AgentSquare fails Tool Creation}: while it optimizes the \textit{mechanism} of tool usage (e.g., choosing between ReAct or Plan-and-Solve), it relies on orchestrating existing tools rather than generating new executable tool implementations from scratch to extend capabilities.

    \item \textbf{ANN}~\citep{ann} conceptualizes multi-agent systems as neural networks, treating agents as learnable nodes and their communication as edges. \textbf{ANN satisfies Optimization}: drawing inspiration from backpropagation, it introduces a ``textual backpropagation'' mechanism that computes textual gradients based on error feedback to iteratively update the agents' system prompts (which function as learnable weights). However, \textbf{ANN fails Agent Creation and Tool Creation}: the framework operates on pre-defined architectural topologies (e.g., Chain, Stack, or Grid structures) with a fixed number of agent nodes; it optimizes the \textit{behavior} of these existing agents rather than autonomously synthesizing new agent roles or executable tools from scratch. Furthermore, it fails \textbf{Experience Persistence}: similar to traditional model training, the learned experience is implicitly internalized into the optimized prompt parameters during the optimization phase. It lacks a dynamic, explicitly retrievable memory bank to persistently store reasoning trajectories, limiting its ability to support lifelong learning or transfer insights to entirely new domains without re-training. Consequently, it also fails \textbf{Experience-Guided Creation}.

    \item \textbf{Alita}~\citep{alita} introduces a self-evolving framework that enables an LLM agent to dynamically expand its capability boundaries by creating and integrating new tools. \textbf{Alita satisfies Tool Creation}: it employs a ``Creator'' module that autonomously synthesizes executable Python tools from scratch to address tasks where existing tools are insufficient. It also satisfies \textbf{Optimization} and \textbf{Experience Persistence}: the framework maintains an explicit ``Experience Pool'' of successful tool-use trajectories and utilizes a ``Promptist'' module to retrieve relevant demonstrations and refine the agent's prompts based on execution feedback. However, \textbf{Alita fails Agent Creation}: it operates as a single-agent system that evolves its tool library, rather than synthesizing new agent roles or collaborative teams. Furthermore, it fails \textbf{Experience-Guided Creation} (in the context of asset generation): while it uses experience to optimize \textit{usage} prompts, the generation of \textit{new tools} is driven by immediate task failures and reflection, without leveraging a retrieval mechanism over historical creation artifacts to guide the synthesis process.

    \item \textbf{ToolMaker}~\citep{toolmaker} introduces a dual-LLM framework where a ``Tool Maker'' autonomously generates reusable Python tools (functions) to address specific tasks, which are then utilized by a ``Tool User'' for subsequent problem-solving. \textbf{ToolMaker satisfies Tool Creation}: its core mechanism is the synthesis of executable code to encapsulate reasoning steps into reusable tools, thereby explicitly expanding the agent's action space. It also satisfies \textbf{Optimization}: during the tool creation phase, it employs a verification loop (utilizing unit tests) to validate the generated code and uses error feedback to iteratively refine and debug the tool until it functions correctly. However, \textbf{ToolMaker fails Agent Creation}: the framework operates with fixed, pre-defined roles (Maker and User) rather than synthesizing new agent personas or collaborative team structures from scratch. Furthermore, it fails \textbf{Experience Persistence} (in the context of cross-task learning) and \textbf{Experience-Guided Creation}: while the generated tools are stored for reuse on instances of the \textit{same} task, the framework does not maintain a retrievable memory bank of creation strategies or past artifacts to guide the generation process for entirely \textit{new, unseen} tasks, effectively facing the cold-start problem for each new domain.

    \item \textbf{AgentVerse}~\citep{agentverse} proposes a flexible multi-agent framework that orchestrates the problem-solving process through four key stages: Expert Recruitment, Collaborative Decision Making, Action Execution, and Evaluation. \textbf{AgentVerse satisfies Agent Creation}: utilizing the Expert Recruitment mechanism, the framework autonomously generates and customizes new agent roles and descriptions tailored to the specific progress of the task, rather than relying on a fixed set of pre-defined personas. It also satisfies \textbf{Optimization}: the Evaluation stage provides real-time feedback on the agents' outcomes, which is used to iteratively refine the collaborative decision-making process and adjust the team's composition or strategies during runtime. However, \textbf{AgentVerse fails Tool Creation}: while agents can utilize existing tools or execute code, the framework focuses on evolving the \textit{team structure} and \textit{agent personas} rather than synthesizing new, reusable executable tool definitions from scratch to expand the action space. Furthermore, it fails \textbf{Experience Persistence} and \textbf{Experience-Guided Creation}: the optimization is confined to the immediate context of the current task loop; it lacks a long-term, retrievable memory mechanism to store successful collaboration patterns or reasoning trajectories for future cross-task transfer, meaning each new task session effectively starts without historical guidance.

    \item \textbf{AutoAgents}~\citep{autoagents} introduces an automatic agent generation framework that dynamically synthesizes a team of specialized agents (including roles and expert profiles) tailored to the specific input task. \textbf{AutoAgents satisfies Agent Creation}: it leverages a ``Planner'' to decompose the task and autonomously generate the identities and descriptions of the necessary expert agents, rather than selecting from a pre-defined library. It also satisfies \textbf{Optimization}: it employs an ``Observer'' mechanism (Agent Observer and Plan Observer) to review the generated agents and execution plans, providing feedback to refine and optimize the team structure and workflows before execution. However, \textbf{AutoAgents fails Tool Creation}: while the generated agents can utilize existing tools, the framework focuses on synthesizing the \textit{agents} themselves, not generating new executable tool code from scratch to expand the system's capabilities. Furthermore, it fails \textbf{Experience Persistence} and \textbf{Experience-Guided Creation}: the generation process is effectively zero-shot for each new task instance; it does not maintain a long-term, retrievable memory of past successful agent configurations or planning trajectories to guide the generation of future agents, tackling each task as an isolated event without accumulation of experience.

    \item \textbf{SwarmAgentic}~\citep{swarmagentic} applies Swarm Intelligence (SI) principles (specifically Particle Swarm Optimization) to the domain of agent system design, treating agents and tools as particles that evolve in a search space. \textbf{SwarmAgentic satisfies Agent Creation and Tool Creation}: the framework autonomously synthesizes both the agent definitions (roles, prompts) and executable tool code (Python functions) from scratch to construct a functional system, rather than selecting from a fixed library. It also satisfies \textbf{Optimization}: it utilizes a velocity-based update mechanism to iteratively refine the agents' prompts and tools based on the ``personal best'' and ``global best'' feedback found during the swarm search process. However, \textbf{SwarmAgentic fails Experience Persistence}: the optimization history and learned patterns are transient, utilized only to converge on a solution for the current task instance. It does not maintain a persistent, retrievable memory bank of successful designs to support lifelong learning across different tasks. Consequently, it also fails \textbf{Experience-Guided Creation}, as the initialization of new systems for unseen tasks relies on zero-shot generation rather than being informed by a repository of historical assets.
\end{itemize}

\appsection{Task Planning} \label{appendix:tool_creation}
In real-world settings, tasks are typically accomplished through multiple steps. To improve the specialization of subsequently created tools and to fully leverage expert agents by assigning them distinct roles aligned with their respective strengths, the system first decomposes a complex task into a set of sub-tasks during the planning stage. Each sub-task specifies its objective, the expected output format, and its dependencies on other sub-tasks—namely, the results from prerequisite sub-tasks required for execution.

\appsection{Tool Creation}
When a sub-task exceeds the agent’s existing capability scope, the system initiates a tool-creation workflow to expand its capability boundaries. However, synthesizing effective tools based solely on sub-task descriptions proves inadequate. Our empirical analysis identifies three key limitations: (1) brief sub-task descriptions provide insufficient constraints, resulting in unstable and inconsistent tool generation; (2) tools derived solely from the model’s internal, static knowledge often lack practical usability; and (3) the inherent stochasticity of model outputs hinders the reproducibility of successful tool-generation processes and prevents effective reuse of failure cases.

To overcome these challenges, we introduce a three-stage tool synthesis strategy: (1) tool specification generation to formalize functionality and interfaces; (2) tool documentation and experience collection to ground tool usage and accumulate actionable knowledge; and (3) tool implementation to produce reliable and reusable tools.

\begin{figure*}
    \centering
        \begin{lstlisting}[style=json]
[
    {
        "tool_name": "simulate_piston_platform_game",
        "tool_description": "Simulates a specific ping-pong game mechanism involving a ball queue, a limited-capacity platform, and three random pistons with complex ejection/replacement rules. Used to calculate the probability of each ball number being 'ejected by a piston' (winning) versus being 'released' (eliminated).",
        "input_parameters": [
            {
                "name": "num_balls",
                "type": "integer",
                "description": "Total number of balls in the queue, numbered 1 to N.",
                "default": 100
            },
            {
                "name": "num_simulations",
                "type": "integer",
                "description": "Monte Carlo iterations.",
                "default": 100000
            }
        ],
        "output_format": {
            "type": "object",
            "properties": {
                "win_probabilities": {
                    "type": "object",
                    "description": "Mapping of ball number to its probability of being ejected by a piston (Winning)."
                },
                "best_choice": {
                    "type": "integer",
                    "description": "The ball number with the highest win probability."
                }
            }
        },
        "core_logic": [
            "Step 1: Initialize `win_counts` for all balls to 0.",
            "Step 2: Run loop `num_simulations` times.",
            "Step 3: In each sim, initialize a queue `deck` [1..num_balls] and a `platform` holding the first 3 balls.",
            "Step 4: While platform is not empty, randomly select a piston (1, 2, or 3) with equal probability.",
            "Step 5: Apply Piston Rules:",
            "   - If Piston 1: Eject Pos 1 (Win). Move Pos 2->1, Pos 3->2. Refill 1 ball from deck to Pos 3.",
            "   - If Piston 2: Eject Pos 2 (Win). Release Pos 1 (Loss/Die). Move Pos 3->1. Refill 2 balls from deck to Pos 2 & 3.",
            "   - If Piston 3: Eject Pos 3 (Win). Release Pos 1 (Loss/Die). Move Pos 2->1. Refill 2 balls from deck to Pos 2 & 3.",
            "Step 6: If a ball is 'Ejected' (Win), increment its count in `win_counts`. If 'Released', do nothing.",
            "Step 7: Continue until platform and deck are empty.",
            "Step 8: Calculate probabilities = wins / total_simulations."
        ]
    }
]
        \end{lstlisting}
    \caption{\textbf{Specification of the tool for simulating the Piston Platform game.} The specification includes the tool name and description, detailed definitions of input parameters and output formats—where each parameter is characterized by its name, type, description, and default value—as well as the core logic of the tool implementation, guiding subsequent tool creation.}
    \label{fig:tool spec case}
\end{figure*}

\begin{itemize}
    \item \textbf{Tool Spec Generation.} Inspired by specification-driven development paradigms~\citep{github_spec_kit_2025}, we require the model to first generate a formal tool specification before implementing the tool itself. As illustrated in Figure~\ref{fig:tool spec case}, this specification includes the \textit{tool name}, a \textit{concise description}, \textit{input parameters}, \textit{output format}, and \textit{core logic}. The core logic is articulated as a sequence of concrete steps that explicitly describe the tool’s execution process (e.g., “Step 1: validate input compliance”). By introducing this intermediate specification stage, the agent can generate tools by directly adhering to well-defined requirements, thereby improving controllability, consistency, and overall accuracy in the tool creation process.

    \item \textbf{Tool Documentation and Experience Collection.} To further mitigate the limitation of relying solely on the parametric, static knowledge, we incorporate an additional grounding step prior to tool generation. Specifically, the agent leverages a web search tool to retrieve external documentation, such as open-source tool descriptions on GitHub\footnote{\url{https://github.com}} and debugging discussions from Stack Overflow\footnote{\url{https://stackoverflow.com/}}. In parallel, the agent queries its Experience Memory using the generated tool specification to retrieve relevant development references. For example, both a tool for extracting basic YouTube video metadata and a tool for downloading YouTube videos can be grounded in documentation for the open-source utility \textit{yt-dlp}~\citep{ytdlp_2025}. By integrating externally sourced documents with experience-based retrieval, the tool creation process is better grounded in real-world implementations, leading to more practical and reliable tools.

    \item \textbf{Tool Implementation.} With a well-defined tool specification and sufficiently rich tool documentation and implementation experience stored in memory, the system can proceed to generate the corresponding tool. As shown in Code~\ref{code:piston_sim}, we intentionally encapsulate each tool in an MCP-compliant format, ensuring that it can be seamlessly integrated by different LLM backbones. This design enables model-agnostic interoperability and facilitates efficient reuse in subsequent applications.

\end{itemize}

\appsection{Assets Recruitment}
To optimize the utilization of tools and expert agents stored in Asset Memory, Mem$^2$Evolve implements an \textit{Assets Recruitment} phase prior to task execution. This mechanism operates at the granularity of sub-tasks. Let $\mathbf{q}=embedding(s_i)$ denote the embedding of the current sub-task description.

\paragraph{Expert Agent Retrieval} 
We query the agent memory $\mathcal{M}_{agt}$ to find the most proficient expert. The retrieval key for a candidate agent $a_i$ is defined as $\mathbf{h}_{a_i} = embedding(\rho_i \oplus \epsilon_i \oplus \sigma_i)$, derived from its role, expertise, and behavior suggestions. We select the optimal agent $a^*$ by retrieving the Top-1 candidate that exceeds a similarity threshold $\delta$:
\begin{equation}
    a^* = \underset{a_i \in \mathcal{M}_{agt}}{\arg\max} \Big\{ \cos(\mathbf{q}, \mathbf{h}_{a_i}) \mid \cos(\mathbf{q}, \mathbf{h}_{a_i}) > \delta \Big\}
\end{equation}
If the set is empty, a new agent generation process is triggered.

\paragraph{Tool Retrieval} 
Similarly, for tool memory $\mathcal{M}_{tool}$, the key is $\mathbf{h}_{t_j} = embedding(n_j \oplus d_{func, j})$. To balance functional support with context window constraints, we retrieve the Top-$K$ relevant tools to form the available toolset $\mathbb{T}_{avail}$:
\begin{equation}
    \mathbb{T}_{avail} = \operatorname*{Top-K}_{t_j \in \mathcal{M}_{tool}} \Big\{ t_j \mid \cos(\mathbf{q}, \mathbf{h}_{t_j}) > \delta \Big\}
\end{equation}

\begin{figure*}
    \centering
        \begin{lstlisting}[style=json]
{ 
  	"role": "Probability Simulation Analyst",
  	"expertise": "Specializes in stochastic modeling and quantitative analysis to derive probabilities from complex mechanical simulations.",
  	"suggestions": [
 		"Execute multiple simulation runs to ensure statistical significance of the results.",
 		"Aggregate ejection data to calculate the specific probability for each ball.",
 		"Identify the ball with the maximum ejection frequency from the dataset."
  	],
  	"tools": [
 		"simulate_ping_pong_game"
  	]
}
        \end{lstlisting}
    \caption{\textbf{Specification of the probabilistic simulation expert.} The specification defines the expert agent’s role, areas of expertise, suggested strategies or recommendations, and the list of tools available for use during task execution. \hongru{ping pong game in tool?}} 
    \label{fig:expert agent case}
\end{figure*}

\appsection{Execution}
Since expert agents must frequently invoke tools to interact with the external environment and make subsequent decisions based on environmental feedback, we adopt the ReAct~\citep{react} framework, which alternates among \textbf{Think}, \textbf{Action}, and \textbf{Observation} steps. Specifically, we design a standardized ReAct system prompt template with dedicated placeholders for prerequisite results from dependent sub-tasks, the expert role, task-specific suggestions, and the set of available tools. During execution, these placeholders are dynamically populated with agent-specific content for each expert agent. This templated design ensures both the stability of the reasoning–action loop and the extensibility of the framework across different expert roles and task settings.

\clearpage
\section{Experimental Details}
\appsection{Baselines}  \label{appendix:baselines}
In this section, we provide detailed descriptions of the baseline frameworks employed in our evaluation, categorized by their operational paradigms.

\paragraph{Naive Large Language Models}
\begin{itemize}
    \item \textbf{Direct}: This is the most fundamental approach, where the task description is fed directly into the Large Language Model (LLM) without any intermediate reasoning steps or external tools. It serves as a baseline to measure the inherent zero-shot capability of the model.
    
    \item \textbf{CoT}~\citep{cot}: CoT enhances the reasoning capabilities of LLMs by prompting them to generate a series of intermediate reasoning steps before producing the final answer. This method is particularly effective for complex tasks requiring multi-step logic.
    
    \item \textbf{ReAct}~\citep{react}: ReAct synergizes reasoning and acting by allowing the model to generate reasoning traces and task-specific actions (such as web searches) in an interleaved manner. This enables the agent to interact with external environments to retrieve information and update its context dynamically.
    
    \item \textbf{OpenAI Deep Research}: A commercial-grade autonomous research agent developed by OpenAI. It is designed to perform deep, multi-step research tasks by browsing the web, synthesizing information from multiple sources, and generating comprehensive reports, representing the state-of-the-art in proprietary agent systems.
\end{itemize}

\paragraph{Experience-Centric Frameworks}
\begin{itemize}
    \item \textbf{DyLAN}~\citep{dylan}: DyLAN is a framework that models multi-agent collaboration as a Temporal Feed-Forward Network. It introduces a ``Team Optimization'' stage utilizing a backward message-passing mechanism to calculate Agent Importance Scores. By actively identifying high-contribution agents and pruning low-performing ones, DyLAN iteratively optimizes the team's collaboration topology, though it relies on a fixed pool of agents rather than creating new ones.
    
    \item \textbf{EvoAgent}~\citep{evoagent}: EvoAgent applies evolutionary algorithms to multi-agent systems, treating agents as individuals in a population. It employs operations such as crossover and mutation on agent prompts to iteratively evolve their behaviors. This allows the system to discover more effective agent personas and strategies over time without manual prompt engineering.

    \item \textbf{AFlow}~\citep{aflow}: AFlow reformulates agentic system development as a search problem over code-represented workflows. Utilizing Monte Carlo Tree Search (MCTS), it iteratively explores and optimizes the design space of workflow structures and node-level prompts. AFlow autonomously constructs new workflow nodes and connections guided by search history, moving away from manual engineering to find the most effective graph topology for specific tasks.

    \item \textbf{DSPy}~\citep{dspy}: DSPy introduces a programming model that abstracts LM pipelines as text transformation graphs. It employs a compiler with various ``teleprompters'' to automatically refine the pipeline's instructions or fine-tune the underlying language model weights based on metric-driven feedback. While it optimizes the pipeline effectively, it relies on user-defined program structures rather than autonomously synthesizing new agent roles or tools.
\end{itemize}

\paragraph{Capacity-Centric Frameworks (Tool-Generative)}
\begin{itemize}
    \item \textbf{Alita}\footnote{We use the implementation at \url{https://github.com/ryantzr1/OpenAlita} as the official code is unavailable.}~\citep{alita}: Alita is a self-evolving framework designed to dynamically expand an agent's capability boundaries. It features a ``Creator'' module that autonomously synthesizes executable Python tools from scratch to address tasks where existing tools are insufficient. Additionally, Alita utilizes a ``Promptist'' module and an explicit Experience Pool to refine usage prompts based on execution feedback, enabling continuous adaptation.
    
    \item \textbf{ToolMaker}~\citep{toolmaker}: ToolMaker adopts a dual-LLM framework comprising a ``Tool Maker'' and a ``Tool User.'' The Maker autonomously generates reusable Python tools (functions) to encapsulate reasoning steps for specific tasks, while the User applies them for problem-solving. It includes a verification loop with unit tests to ensure the reliability of the generated code, explicitly expanding the agent's action space through code synthesis.
\end{itemize}

\paragraph{Capacity-Centric Frameworks (Agent-Generative)}
\begin{itemize}
    \item \textbf{AgentVerse}~\citep{agentverse}: AgentVerse proposes a flexible multi-agent framework orchestrating the problem-solving process through Expert Recruitment, Collaborative Decision Making, Action Execution, and Evaluation. It autonomously generates and customizes new agent roles tailored to the task progress. The framework uses real-time feedback to refine the collaborative process and adjust team composition during runtime.
    
    \item \textbf{AutoAgents}~\citep{autoagents}: AutoAgents introduces an automatic agent generation framework that dynamically synthesizes a team of specialized agents tailored to the input task. Leveraging a ``Planner'' to decompose tasks, it autonomously generates expert identities and descriptions. An ``Observer'' mechanism further reviews and refines the execution plans and team structure, ensuring the generated agents are optimized for the specific problem context.
    
    \item \textbf{SwarmAgentic}~\citep{swarmagentic}: SwarmAgentic applies Swarm Intelligence principles, specifically Particle Swarm Optimization (PSO), to agent system design. It treats agents and tools as particles evolving in a search space, autonomously synthesizing both agent definitions and executable tool code. The framework utilizes velocity-based updates to iteratively refine prompts and tools based on ``personal best'' and ``global best'' feedback found during the swarm search.
\end{itemize}

\appsection{Benchmarks} \label{appendix:benchmarks}
\begin{table}[!htbp]
    \centering
    \small
    \caption{\textbf{Overview of the benchmarks, domains, test set sizes, and evaluation metrics used in  experiments.} $^{\dagger}$ indicates that the test set was randomly sampled. $^{\ddagger}$Average Score is the mean of Delivery Rate, Micro/Macro Commonsense Constraint Pass Rate, and Micro/Macro Hard Constraint Pass Rate.}
    \label{tab:benchmarks_overview}
    \begin{tabular}{llcl}
        \toprule
        \textbf{Benchmark} & \textbf{Domain} & \textbf{Test Size} & \textbf{Metric} \\
        \midrule
        GAIA~\citep{gaia}& General Assistant & 166 & Pass@1 \\
        ALFWorld~\citep{alfworld} & Embodied Task & 134 & Success Rate \\
        HotpotQA~\citep{hotpotqa} & Multi-hop QA & 500$^{\dagger}$ & Exact Match (EM) \\
        2WikiMultihopQA~\citep{2wiki} & Multi-hop QA & 500$^{\dagger}$ & Exact Match (EM) \\
        AIME 24 & Math Reasoning & 30 & Pass@1 \\
        AIME 25 & Math Reasoning & 30 & Pass@1 \\
        TravelPlanner~\citep{travelplanner} & Complex Planning & 1,000 & Average Score$^{\ddagger}$ \\
        WebShop~\citep{webshop} & Web Navigation & 251 & Success Rate \\
        \bottomrule
    \end{tabular}
\end{table}

To evaluate the general-purpose task-solving capability of DEMA, we conduct experiments across six task categories and eight benchmarks, as summarized in Table~\ref{tab:benchmarks_overview}.

\begin{itemize}
\item \textbf{GAIA}: GAIA is a benchmark designed to assess the capabilities of general-purpose AI assistants. It consists of 466 real-world, scenario-based questions covering daily tasks, scientific reasoning, web browsing, and tool usage. While these tasks are conceptually simple for humans, they remain challenging for advanced AI systems. We report Pass@1 as the primary evaluation metric.

\item \textbf{ALFWorld}: ALFWorld aligns text-based games with embodied environments to evaluate an agent’s ability to reason and act in interactive settings. It requires the agent to understand high-level goals and execute a sequence of low-level actions to interact with objects. We evaluate our method on 134 tasks and use Success Rate as the evaluation metric.

\item \textbf{HotpotQA}: HotpotQA is a question-answering benchmark that challenges agents to perform multi-hop reasoning across multiple documents to locate relevant facts. In our experiments, we randomly sample a subset of 500 instances as the test set. Performance is evaluated using the Exact Match (EM) metric.

\item \textbf{2WikiMultihopQA}: Similar to HotpotQA, 2WikiMultihopQA evaluates multi-hop reasoning capabilities over Wikipedia articles and features complex queries that require synthesizing information from multiple sources. We randomly sample 500 instances for testing and use EM for evaluation.

\item \textbf{AIME 24/25}: These benchmarks correspond to the problem sets from the 2024 and 2025 American Invitational Mathematics Examinations. Each dataset consists of 30 high-difficulty problems that test advanced mathematical reasoning and creative problem-solving abilities. We use Pass@1 to measure accuracy.

\item \textbf{TravelPlanner}: TravelPlanner evaluates language agents in complex tool-use and long-horizon planning scenarios, such as generating travel itineraries under strict constraints. We use the full test set of 1,000 instances. The final score is computed as the average of five metrics: Delivery Rate, Micro Commonsense Constraint Pass Rate, Macro Commonsense Constraint Pass Rate, Micro Hard Constraint Pass Rate, and Macro Hard Constraint Pass Rate.

\item \textbf{WebShop}: WebShop is a simulated e-commerce environment containing over one million real-world products. It tests an agent’s ability to navigate web pages, search, browse, and select options to fulfill user instructions. We evaluate on 251 test instances and measure performance using Success Rate.

\end{itemize}

\appsection{RQ1: Ablation Study} \label{appendix: ablation}
\begin{table*}[!htbp]
\centering
\small
\setlength{\tabcolsep}{2.5pt} 
\renewcommand{\arraystretch}{1.3} 

\resizebox{\textwidth}{!}{%
\begin{tabular}{l c c c c c c c c c}
\toprule[0.08em]
\multirow{2.5}{*}{\textbf{Method}} &
\textbf{GAIA} &
\textbf{Embodied} &
\multicolumn{2}{c}{\textbf{Multi-Hop QA}} &
\multicolumn{2}{c}{\textbf{Math}} &
\textbf{Planning} &
\textbf{Web} &
\multirow{2.5}{*}{\textbf{Avg.}} \\

\cmidrule(lr){2-2}
\cmidrule(lr){3-3}
\cmidrule(lr){4-5}
\cmidrule(lr){6-7}
\cmidrule(lr){8-8}
\cmidrule(lr){9-9}

& \textbf{Total} & \textbf{ALFWorld} & \textbf{HotpotQA} & \textbf{2Wiki} & \textbf{AIME24} & \textbf{AIME25} & \textbf{TravelPlanner} & \textbf{WebShop} & \\

\midrule[0.05em]

\textbf{Mem$^\textbf{2}$Evolve (Ours)} &
\textbf{76.31} & \textbf{94.31} & \textbf{60.80} & \textbf{82.00} & \textbf{76.70} & \textbf{73.33} & \textbf{59.25} & \textbf{39.20} & \textbf{70.24} \\

\midrule[0.05em]
\rowcolor{blue!7}	
\multicolumn{10}{c}{\textit{\textbf{w/o Asset Creation}}} \\
\midrule[0.05em]

w/o Tool Creation & 
21.69 \down{54.62} & 94.10 \down{0.21} & 59.50 \down{1.30} & 81.50 \down{0.50} & 66.67 \down{10.03} & 60.00 \down{13.33} & 57.15 \down{2.10} & 39.05 \down{0.15} & 59.96 \down{10.28} \\

w/o Expert Agent Creation & 
75.30 \down{1.01} & 93.65 \down{0.66} & 59.00 \down{1.80} & 81.00 \down{1.00} & 73.33 \down{3.37} & 70.00 \down{3.33} & 57.08 \down{2.17} & 38.80 \down{0.40} & 68.52 \down{1.72} \\

\midrule[0.05em]
\rowcolor{blue!7}	
\multicolumn{10}{c}{\textit{\textbf{w/o Experience Distillation}}} \\
\midrule[0.05em]

w/o Tool Memory & 
67.47 \down{8.84} & 94.25 \down{0.06} & 60.00 \down{0.80} & 81.50 \down{0.50} & 70.00 \down{6.70} & 66.67 \down{6.66} & 57.85 \down{1.40} & 39.15 \down{0.05} & 67.11 \down{3.13} \\

w/o Agent Memory & 
74.70 \down{1.61} & 88.06 \down{6.25} & 56.80 \down{4.00} & 77.40 \down{4.60} & 73.33 \down{3.37} & 70.00 \down{3.33} & 49.36 \down{9.89} & 34.40 \down{4.80} & 65.51 \down{4.73} \\

\bottomrule[0.08em]
\end{tabular}%
}
\caption{\textbf{Ablation study of Mem$^\textbf{2}$Evolve}. Pass@1 scores are reported. Performance drops \textcolor{downred}{(\textbf{$\downarrow$})} relative to the full model are shown in parentheses.}
\label{tab:ablation_results}
\end{table*}

\clearpage
\section{Prompt Template}
\phantomsection                      
\refstepcounter{subsection}          
\addcontentsline{toc}{subsection}{\protect\numberline{\thesubsection}Task Planning} 

\begin{mybox}{Prompt for Task Planning}
\begin{lstlisting}[style=promptstyle]
(*\textcolor{blue}{You are a task planning expert. Analyze the query carefully and break down complex queries into logical, actionable steps.}*)

USER QUERY: (*\textcolor{blue}{\{query\}}*)

You MUST output EXACTLY ONE JSON object with this structure:
{
  "sub_tasks": [
    {
      "description": "Specific actionable task...",
      "dependencies": []
    },
    {
      "description": "Specific actionable task...",
      "dependencies": []
    }
  ]
}
RULES:
1. Each task description should be COMPLETE and SPECIFIC
2. Use task numbers (1,2,3...) for dependencies
3. First task usually has no dependencies: "dependencies": []
4. If task 2 depends on task 1: "dependencies": [1]
5. If task 3 depends on tasks 1 and 2: "dependencies": [1, 2]
6. Preserve the EXACT units and format requirements from the original query in task descriptions
7. Output pure JSON format with no other content.
\end{lstlisting}
\end{mybox}
\label{prompt: task planning}

\bigskip

\phantomsection                      
\refstepcounter{subsection}          
\addcontentsline{toc}{subsection}{\protect\numberline{\thesubsection}Assess Tool Need}

\begin{mybox}{Prompt for Assess Tool Need}
\begin{lstlisting}[style=promptstyle]
You are a Principal Engineer. Your goal is to analyze the given multi-step task plan and determine the most efficient way to solve it. (*\textcolor{blue}{Your primary objective is to use LLM-Native capabilities and reuse existing tools. You will only determine that new tools are needed (need\_creation: true) as a last resort.}*)


- Core Principles for Analysis
  - Maximize LLM-Native Capabilities (Highest Priority): You (the LLM) are the primary tool. You can perform many tasks natively without code. Do NOT propose a tool for any step that involves:
    - Reasoning, planning, or making decisions based on provided context.
    - Extracting, reformatting, filtering, sorting, or summarizing data from context.
    - Simple data transformations or list operations (e.g., finding an item, counting).
    - Any task that doesn't strictly require one of the "Why Code?" principles below.
  - Reuse First, Create Last (Second Priority): After exhausting LLM-Native capabilities, your next goal is to reuse available_tools. Do NOT propose duplicates or near-duplicates.
  - Analyze the Full Plan: Review the entire sequence of steps to understand overall goals and data flow.
  - Deconstruct Each Step: Mentally decompose each step into atomic required_capabilities.
  - Justify the Need for a Tool (Why Code?): A step requires a code tool only if it involves at least one of:
    - Complex Calculations: Non-trivial math, statistics, physics (beyond simple arithmetic).
    - Stateful Simulation/Iteration: Executing a loop many times while tracking changing state.
    - Complex Data Manipulation: Creating/modifying nested data structures according to precise rules (not simple extraction).
    - External I/O: Accessing APIs, files, databases, or system resources.
    - Deterministic Logic: When a precise, non-negotiable output is required based on complex rules.

- Decision Procedure
  - Decompose Plan: Break down the plan into a flat list of atomic required_capabilities.
  - Evaluate Each Capability (3-step check):
      a. LLM-Native? Can this capability be handled directly by the LLM (based on Maximize LLM-Native Capabilities)? If yes - it's covered.
      b. Existing Tool? If not LLM-native - can it be handled by one or more available_tools? If yes - add those tool(s) to matching_tools.
      c. Gap Identified? Only if both above answers are NO - this capability is truly missing/gap.

- Determine Final Output
      If all capabilities are covered by either above method:
        need_creation=false; missing_tools=[]
      Else if any essential gaps remain from above analysis:
        need_creation=true; propose minimal set of missing_tools covering ONLY those gaps

STRICT TYPE CONTRACT
- For every field listed below, emit values with the exact type specified.
- If you don't know a value, use an empty array [] or null where appropriate, never change the type.
- Constraints per field (types in parentheses):
  - required_capabilities (string[]): flat list of atomic capabilities.
  - matching_tools (string[]): names of existing tools.
  - missing_tools (object[]): each item MUST be an object with fields:
    - name (string)
    - description (string)
    - reason (string)
    - example_input_output (string) - a single-line string formatted as: "Input: {...} -> Output: {...}".
  - justification (string)
  - need_creation (boolean) - Phase 1 only: whether new tools are required.

Output Format

{
  "need_creation": true | false,
  "required_capabilities": ["verb_object", "..."],
  "matching_tools": ["existing_tool_a", "existing_tool_b"],
  "missing_tools": [
    {
      "name": "tool_for_gap",
      "description": "Does exactly this: ...",
      "reason": "Essential because: [Why Code? principle justification]",
      "example_input_output": "Input: {\"x\": 1} -> Output: {\"y\": 2}"
    }
  ],
  "justification": "1-3 sentences explaining overall coverage and gaps."
}

Validation Checklist
- matching_tools in current available tool names (exact match).
- missing_tools names are unique and non-empty; reasons reference the Why Code? principles.
- If existing tools fully cover all required_capabilities: need_creation=false and missing_tools=[].
- Output must be a single JSON object with no extra commentary.
- Each tool does exactly one specific operation.

EXAMPLES OF GOOD TOOLS:
- `count_even_numbers` - counts even numbers in a list
- `extract_text_from_image` - extracts text from image (uses built-in vision)
- `calculate_percentage` - calculates percentage from two numbers
- `parse_addresses` - extracts and processes address data

EXAMPLES OF BAD TOOLS:
- `address_parser` + `parity_checker` + `sunset_awning_counter` (redundant!)
- `complex_data_analyzer` (too vague)
- `multi_purpose_processor` (does too many things)

Task:
{task}

Current Available Tools:
{available_tools}
\end{lstlisting}
\end{mybox}

\bigskip

\phantomsection                      
\refstepcounter{subsection}          
\addcontentsline{toc}{subsection}{\protect\numberline{\thesubsection}Tool Spec Creation}

\begin{mybox}{Prompt for Tool Spec Creation}
\begin{lstlisting}[style=promptstyle]
(*\textcolor{blue}{You are a Principal Engineer responsible for writing implementable specifications for new tools. Use the brief descriptions to generate detailed creation specs for ONLY those tools. }*)

Design Principles (reused and specialized)
- One spec per missing tool; keep names consistent with Phase 1.
- Avoid overlapping with Current Available Tools; if overlap is detected, omit that spec.

STRICT TYPE CONTRACT
- For every field listed below, emit values with the exact type specified.
- If you don't know a value, use an empty array [] or null where appropriate, never change the type.

Output Format:
{
  (*\textcolor{blue}{"tool\_creation\_specs"}*): [
    {
      (*\textcolor{blue}{"tool\_name"}*): "tool_for_gap",
      (*\textcolor{blue}{"tool\_description"}*): "1-2 sentence summary.",
      (*\textcolor{blue}{"input\_parameters"}*): [ 
          { "name": "param", "type": "string", "description": "...", "required": true } 
      ],
      (*\textcolor{blue}{"output\_format"}*): { "type": "object", "properties": { "result": { "type": "number" } } },
      (*\textcolor{blue}{"core\_logic"}*): [
        "Step 1: Input validation ...",
        "Step 2: Core processing ..."
      ]
    }
  ]
}

Task:
(*\textcolor{blue}{\{task\}}*)

Current Available Tools:
(*\textcolor{blue}{\{available\_tools\}}*)
\end{lstlisting}
\end{mybox}

\bigskip

\phantomsection                      
\refstepcounter{subsection}          
\addcontentsline{toc}{subsection}{\protect\numberline{\thesubsection}Tool Creation} 

\begin{mybox}{Prompt for Tool Creation}
\begin{lstlisting}[style=promptstyle]
You are a senior software engineering specialist building robust, reusable MCP (Model Context Protocol) tools following Claude Desktop standards.

OBJECTIVE
Create a powerful and high-quality MCP tool to accurately and effectively solve the tasks specified below. This tool should be a direct and effective solution to a given problem, and it should be generated strictly according to the **Core Implementation Logic**.

TASK
(*\textcolor{blue}{\{original\_query\}}*)

TOOL NAME
(*\textcolor{blue}{\{tool\_name\}}*)

Tool Description:
(*\textcolor{blue}{\{tool\_description\}}*)

Core Implementation Logic:
(*\textcolor{blue}{\{core\_logic\}}*)

Input Parameters:
(*\textcolor{blue}{\{input\_parameters\}}*)

Output Format:
(*\textcolor{blue}{\{output\_format\}}*)


DESIGN PRINCIPIPLES
- Solve the Problem Precisely: The tool's core goal is to provide a direct and effective solution for the current task. Prioritize functional correctness and robustness over premature generalization.
- Clear Interface Design: Clearly define the tool's input parameters and output format. Even if the tool has a specific function, its interface should be clear, well-documented, and easy to use.
- Professional Naming Convention: Choose a descriptive, domain-prefixed snake_case name that accurately reflects the tool's specific purpose.

OUTPUT CONTRACT (STRICT)
Return EXACTLY ONE JSON object as plain text (no markdown fences/backticks or extra prose). The object must have these keys:

1. name: string - professional general-purpose snake_case with domain prefix
   - Must be EXACTLY "(*\textcolor{blue}{\{tool\_name\}}*)" (do not rename or vary)

2. description: string - DETAILED description (5-10 sentences) covering:
   - What the tool does (core functionality)
   - Key capabilities and features
   - Typical use cases and scenarios
   - Input/output data types and formats
   - Any limitations or constraints

3. input_schema: object - JSON Schema (Claude Desktop standard) defining parameters:
   {{
     "type": "object",
     "properties": {{
       "param_name": {{
         "type": "string|number|boolean|array|object",
         "description": "Detailed parameter description",
         "enum": ["optional", "allowed", "values"],
         "default": "optional_default_value",
         "example": "input example"
       }}
     }},
     "required": ["list_of_required_params"]
   }}
   - Include ALL parameters the tool accepts
   - Provide detailed descriptions for each parameter
   - Specify types, constraints, enums, and defaults
   - Mark required vs optional parameters clearly

4. returns: object - Output format specification:
   {{
     "type": "string|object",
     "description": "Detailed description of return value",
     "format": "json|text|structured",
     "schema": {{"optional": "output schema for structured returns"}}
   }}

5. module_code: string - COMPLETE Python module source to be saved as `<tool_name>.py`

Module requirements (STRICT):
- Include necessary imports.
- Define ONE public function implementing the tool with an EXPLICIT parameter list derived from input_schema (no **kwargs).
- Implement robust validation and error handling per input_schema (types, required fields, enums, ranges).
- Provide clear control flow and helpful error messages referencing parameter names.
- Define TOOL_CONFIG = {{ "name": "(*\textcolor{blue}{\{tool\_name\}}*)", "description": <desc>, "function": <function_object>, "input_schema": <schema>, "returns": <returns> }} at module scope.
- Multi-line Python code with normal 4-space indentation.
- No external network calls or dependencies unless they are clearly documented and optional.

Naming constraints (STRICT):
- The JSON field name MUST equal "(*\textcolor{blue}{\{tool\_name\}}*)".
- In the Python module, TOOL_CONFIG['name'] MUST equal "(*\textcolor{blue}{\{tool\_name\}}*)".
- Assume the file will be saved as "(*\textcolor{blue}{\{tool\_name\}}*)".py; do not reference any other module name.

Formatting constraints:
- No markdown fences.
- No trailing backslashes at line ends.
- The module must be self-contained and importable.

CAPABILITIES & QUALITY
- Robust Input Handling: The tool must robustly handle various expected inputs and provide clear, helpful error messages for invalid or edge-case inputs.
- Graceful Degradation: Provide graceful degradation when inputs or external resources (if any) are missing or invalid.
- Strict Validation: Validate all inputs against the input_schema at the beginning of the function.
- Code Excellence: Optimize for clarity, maintainability, and performance; follow PEP 8 standards.
- Helpful Errors: Add helpful error messages that reference parameter names to guide the user.

REMINDERS
- Output must be a single JSON object (no additional commentary).
- Absolutely no json or python fences in the output.
- Ensure input_schema follows JSON Schema specification exactly.
- Provide detailed, production-ready documentation in all fields.

\end{lstlisting}
\end{mybox}
\label{prompt: crete tool}

\bigskip

\phantomsection                      
\refstepcounter{subsection}          
\addcontentsline{toc}{subsection}{\protect\numberline{\thesubsection}Agent Creation} 

\begin{mybox}{Prompt for Agent Creation}
\begin{lstlisting}[style=promptstyle]
You are an expert agent designer. Create a minimal, reusable specialist agent specification for the task.

Inputs
- User Task: (*\textcolor{blue}{\{sub\_task\}}*)
- Available Tools: (*\textcolor{blue}{\{tools\}}*)

Output format
- Return exactly one JSON object as the entire response.
- The object must contain only these four keys, in this exact order: role, suggestions, tools, expertise.
- Explicit schema:
  - role: string
  - suggestions: array of 3-5 strings
  - tools: array of 0-3 strings (tool names from Available Tools)
  - expertise: string (1-2 sentences; concise domain strengths and typical methodology)
- Do not include any other text, comments, code fences, or fields.

Constraints
Role
- A clear, human-readable professional title suitable for a business card.
- 2-5 words, Title Case, English only, no emojis/symbols.

Suggestions
- An array of 3-5 short, concrete execution hints for approaching this class of tasks.
- Each item starts with an imperative verb and is 6-16 words.
- Specific within the domain implied by (*\textcolor{blue}{\{sub\_task\}}*), yet general-purpose (not tailored to a single query).
- No placeholders, no meta references (e.g., "JSON", "this prompt"), avoid tool names unless essential to the method.

Tools
- Choose a minimal subset of tool names taken verbatim from Available Tools.
- Use exact casing/spelling; do not invent, modify, or describe tools; no duplicates.
- If no tool is applicable or Available Tools is empty, use an empty array [].

Expertise
- 1-2 sentences (20-160 characters) summarizing domain strengths and typical methodology.
- Mention information gathering, evaluation, analysis (quantitative/qualitative as applicable), and synthesis.
- Avoid listing tools; focus on capabilities and working approach.

General
- Ensure the role, suggestions, tools, and expertise are coherent with one another and with (*\textcolor{blue}{\{sub\_task\}}*).
- Do not echo the inputs. Do not add explanations.
- Output must be valid JSON using double quotes and no trailing commas.
\end{lstlisting}
\end{mybox}
\label{prompt: crete tool}

\bigskip

\phantomsection                      
\refstepcounter{subsection}          
\addcontentsline{toc}{subsection}{\protect\numberline{\thesubsection}React Template}

\begin{mybox}{Prompt for React Template}
\begin{lstlisting}[style=promptstyle]
You are a (*\textcolor{blue}{\{role\}}*). Based on prior agents' results and completed steps, your goal is to complete the current task as effectively as possible.

# Suggestions
(*\textcolor{blue}{\{suggestions\}}*)

# Progress So Far
(*\textcolor{blue}{\{previous\}}*)

# Past Experience References
Below are relevant experiences from previous similar tasks. Reference the success experiences to learn effective approaches, and learn from the failure experiences to avoid common pitfalls.

## Success Experiences
(*\textcolor{blue}{\{success\_experiences\}}*)

## Failure Experiences
(*\textcolor{blue}{\{failure\_experiences\}}*)

You have access to the following tools:
# Tools
(*\textcolor{blue}{\{tool\}}*)

# Steps
1. Review the outputs from previous agents to understand progress and context.
2. Analyze the task and decompose it if needed. Utilize the available tools as appropriate.
3. Define the current step you will complete, labeling it as 'CurrentStep'.
4. Choose one Action from the available tools to execute the current step.

# Format example
(*\textcolor{blue}{\{format\_example\}}*)
\end{lstlisting}
\end{mybox}
\label{prompt: crete tool}

\bigskip

\phantomsection                      
\refstepcounter{subsection}          
\addcontentsline{toc}{subsection}{\protect\numberline{\thesubsection}LLM as a Judge} 

\begin{mybox}{Prompt for LLM as a Judge}
\begin{lstlisting}[style=promptstyle]
You are a Judge LLM responsible for evaluating the entire execution trajectory of a task and determining whether it was completed correctly.

OBJECTIVE
Analyze the complete task execution trajectory and provide a comprehensive evaluation of task completion, agent performance, and tool effectiveness.

TRAJECTORY INFORMATION

Original Query:
(*\textcolor{blue}{\{query\}}*)

Decomposed Subtasks:
(*\textcolor{blue}{\{subtasks\}}*)

Agent Assignments:
(*\textcolor{blue}{\{agent\_assignments\}}*)

Agent Execution Processes:
(*\textcolor{blue}{\{agent\_processes\}}*)

Result Aggregation Process:
(*\textcolor{blue}{\{aggregation\_process\}}*)

Final Result:
(*\textcolor{blue}{\{final\_result\}}*)

Newly Created Tools (if any):
(*\textcolor{blue}{\{created\_tools\}}*)

EVALUATION CRITERIA

1. Task Completion Assessment:
   - Does the final result correctly answer the original query?
   - Are all decomposed subtasks properly addressed?
   - Is the aggregated result logically coherent and complete?

2. Agent Performance Assessment:
   - Did each agent execute its assigned subtask correctly?
   - Were the agents' reasoning processes sound and effective?
   - Did agents properly utilize available tools?

3. Tool Effectiveness Assessment:
   - Did existing tools function correctly when used?
   - Were newly created tools (if any) implemented correctly?
   - Did tools produce expected outputs?

EVALUATION OUTPUT FORMAT

You must provide your evaluation in the following JSON format:

{{
  "task_completed": true/false,
  "completion_quality": "good/poor",
  "overall_assessment": "Detailed overall assessment of task completion",
  "agent_evaluations": [
    {{
      "agent_id": "agent identifier",
      "agent_role": "role name",
      "subtask_id": "subtask identifier",
      "performance": "success/failure",
      "strengths": ["strength 1", "strength 2", ...],
      "issues": ["issue 1", "issue 2", ...]
      }},
    ...
  ],
  "tool_evaluations": [
    {{
      "tool_name": "tool name",
      "tool_type": "existing/newly_created",
      "effectiveness": "success/partial_success/failure",
      "issues": ["issue 1", "issue 2", ...],
    }},
    ...
  ]
}}

IMPORTANT GUIDELINES

1. Be objective and thorough in your evaluation
2. Provide specific, actionable feedback
3. Identify both strengths and weaknesses
4. Focus on patterns that can inform future improvements
5. Distinguish between agent failures and tool failures
6. Consider the complexity of the task when evaluating performance
7. Ensure all JSON is properly formatted and valid

Return ONLY the JSON evaluation, no additional text.
\end{lstlisting}
\end{mybox}
\label{prompt: crete tool}

\bigskip

\phantomsection                      
\refstepcounter{subsection}          
\addcontentsline{toc}{subsection}{\protect\numberline{\thesubsection}Tool Memory Generation}

\begin{mybox}{Prompt for Tool Memory Generation}
\begin{lstlisting}[style=promptstyle]
You are a technical documentation specialist creating comprehensive tool implementation guides.

OBJECTIVE
Generate a detailed markdown section that captures the implementation experience of this tool for future reference and reuse. This will be one entry within a broader topic file.

TOOL INFORMATION
Tool Name: (*\textcolor{blue}{\{tool\_name\}}*)
Tool Description: (*\textcolor{blue}{\{tool\_description\}}*)
Required Capabilities: (*\textcolor{blue}{\{required\_capabilities\}}*)
Usage Examples: (*\textcolor{blue}{\{usage\_examples\}}*)

Tool Implementation Code:
(*\textcolor{blue}{\{tool\_code\}}*)

DOCUMENT STRUCTURE
Create a markdown section with the following structure:

## [Generate a specific "How to..." question title here]

The title should:
- Start with "How to" or "How can I"
- Be specific to what this exact implementation does
- Be 5-15 words long
- Clearly distinguish this implementation from similar ones
- Example: "How to Download Files From a URL?", "How to Parse CSV Files with Custom Delimiters?"

### Description
A concise 2-3 sentence overview explaining what this specific implementation does and what problem it solves.

### Use Cases
A bulleted list of practical scenarios where this implementation is applicable. Include:
- Specific real-world situations
- Types of data or inputs it handles
- Common integration patterns

### Code Implementation
The core implementation of the tool. Format as:
```python
[Include the complete, clean tool implementation code here]
```

### Tool Configuration
(Optional - only include if the tool uses API keys, URLs, or other configuration)
Document any configuration requirements:
- API keys needed
- Environment variables
- URL endpoints
- Configuration parameters

QUALITY REQUIREMENTS
1. Write in clear, professional technical documentation style
2. Use proper markdown formatting (## for main title, ### for subsections)
3. Be concise but comprehensive
4. Focus on reusability and understanding
5. Include practical context, not just code
6. Highlight key implementation patterns
7. Note any important dependencies or requirements

OUTPUT FORMAT
Return ONLY the complete markdown section. Do not include any preamble, explanations, or commentary outside the document itself. The section MUST start with "## How to..." as the first line.

\end{lstlisting}
\end{mybox}
\label{prompt:tool memory}

\bigskip

\phantomsection                      
\refstepcounter{subsection}          
\addcontentsline{toc}{subsection}{\protect\numberline{\thesubsection}Success Agent Memory Generation} 

\begin{mybox}{Prompt for Success Agent Memory Generation}
\begin{lstlisting}[style=promptstyle]
You are an experience synthesis specialist creating memory entries for an agent based on successful task execution.

OBJECTIVE
Generate a comprehensive memory entry that captures the successful experience of this agent, including successful strategies, effective tool usage, and valuable insights for future similar tasks.

Agent Role: (*\textcolor{blue}{\{agent\_role\}}*)
Agent Skills: (*\textcolor{blue}{\{agent\_skills\}}*)

TASK EXECUTION INFORMATION
Subtask Description: (*\textcolor{blue}{\{subtask\_description\}}*)
Task Context: (*\textcolor{blue}{\{task\_context\}}*)
Tools Used: (*\textcolor{blue}{\{tools\_used\}}*)
Execution Process:
(*\textcolor{blue}{\{execution\_process\}}*)

Final Result: (*\textcolor{blue}{\{final\_result\}}*)

JUDGE EVALUATION
Performance Rating: (*\textcolor{blue}{\{performance\_rating\}}*)
Strengths Identified: (*\textcolor{blue}{\{strengths\}}*)
Key Success Patterns: (*\textcolor{blue}{\{success\_patterns\}}*)

MEMORY ENTRY STRUCTURE

Generate a memory entry in the following markdown format:

## [Create a specific, descriptive title for this experience]

The title should:
- Be specific to the type of task handled
- Highlight the key capability demonstrated
- Be 5-15 words long
- Example: "How to Extract and Summarize Information from Multiple Web Sources"

### Description
A concise 2-3 sentence overview of what this experience teaches about handling this type of task.

### Task Context
Describe the original problem and the conditions under which this task was performed.

### Experience of Success
Detail the successful strategy and reasoning process:
- **Successful Strategy**: Key decision points, problem-solving approach, and how tools were selected
- **Tools and Techniques**: Tools used and how they contributed to success
- **Key Insights**: What made this approach effective and when to apply similar strategies
- **Applicable Scenarios**: Similar task types where this experience would be valuable
- **Potential Pitfalls**: Conditions that might cause similar approaches to fail and warning signs to watch for

QUALITY REQUIREMENTS
1. Write in clear, professional documentation style
2. Use proper markdown formatting
3. Be specific and actionable
4. Focus on transferable knowledge
5. Highlight decision-making processes
6. Include both what worked and why it worked

OUTPUT FORMAT
Return ONLY the complete markdown memory entry. Do not include any preamble or explanations outside the memory entry itself. The entry MUST start with "##" as the first line.
\end{lstlisting}
\end{mybox}
\label{prompt:agent_memory}

\bigskip

\phantomsection                      
\refstepcounter{subsection}          
\addcontentsline{toc}{subsection}{\protect\numberline{\thesubsection}Failure Agent Memory Generation} 

\begin{mybox}{Prompt for Failure Agent Memory Generation}
\begin{lstlisting}[style=promptstyle]
You are an error analysis specialist creating memory entries that help agents learn from failures.

OBJECTIVE
Generate a comprehensive memory entry that captures the failure experience, analyzes root causes, and provides clear guidance on how to avoid similar failures in the future.

AGENT INFORMATION
Agent Role: (*\textcolor{blue}{\{agent\_role\}}*)
Agent Skills: (*\textcolor{blue}{\{agent\_skills\}}*)

TASK EXECUTION INFORMATION
Subtask Description: (*\textcolor{blue}{\{subtask\_description\}}*)
Task Context: (*\textcolor{blue}{\{task\_context\}}*)
Tools Attempted: (*\textcolor{blue}{\{tools\_used\}}*)
Execution Process:
(*\textcolor{blue}{\{execution\_process\}}*)

Failure Outcome: (*\textcolor{blue}{\{failure\_outcome\}}*)

JUDGE EVALUATION
Performance Rating: (*\textcolor{blue}{\{performance\_rating\}}*)
Issues Identified: (*\textcolor{blue}{\{issues\}}*)
Root Cause Analysis: (*\textcolor{blue}{\{root\_cause\}}*)
Suggested Fixes: (*\textcolor{blue}{\{suggested\_fixes\}}*)

MEMORY ENTRY STRUCTURE

Generate a memory entry in the following markdown format:

## [Create a specific, descriptive title for this failure case]

The title should:
- Clearly indicate the problematic scenario
- Be specific enough to match similar future situations
- Be 5-15 words long
- Example: "Common Pitfalls When Parsing Inconsistent CSV Data Formats"

### Description
A concise 2-3 sentence overview of what went wrong and the key lesson to take away.

### Task Context
Describe the task and conditions that led to this execution:
- What was attempted
- What was expected
- What assumptions were made

### Experience of Failure
Detail what went wrong and how to avoid it:
- **What Went Wrong**: Specific errors, incorrect decisions, tools misused, or logic errors
- **Root Cause Analysis**: Why the approach didn't work, what was misunderstood, gaps in knowledge
- **Corrective Actions**: Alternative approaches, additional checks, tools to use instead, validation steps
- **Warning Signs**: Task characteristics that trigger this issue, context patterns, red flags during execution
- **Related Success Patterns**: Successful approaches that should be used instead
- **Partial Successes**: Parts of the approach that were correct, tools or techniques that functioned properly

QUALITY REQUIREMENTS
1. Write in clear, instructive style
2. Use proper markdown formatting
3. Be honest and specific about failures
4. Provide actionable corrective guidance
5. Focus on learning and improvement
6. Help prevent similar failures

OUTPUT FORMAT
Return ONLY the complete markdown memory entry. Do not include any preamble or explanations outside the memory entry itself. The entry MUST start with "##" as the first line.

\end{lstlisting}
\end{mybox}

\clearpage

\section{Case Study} \label{appendix: case_rq}
\appsection{Tool Implementation for \textit{Simulate Piston Platform Game}}
As shown in Code~\ref{code:piston_sim}, when performing complex probabilistic reasoning tasks, the agent extends its modeling capabilities by constructing a simulation tool and employing Monte Carlo methods to estimate the true underlying probabilities. The implementation of this tool is entirely grounded in the original problem formulation and faithfully simulates all rules specified in the task. Moreover, the implementation exhibits strong robustness, supporting simulations with arbitrary numbers of trials and arbitrary numbers of balls, and is, to some extent, adaptable to multiple variants of the original problem.

\begin{lstlisting}[style=pythonstyle, caption={Code Implenmenation for Simulate Piston Platform Game Tool}, label={code:piston_sim}]
import random

def simulate_piston_platform_game(num_balls=100, num_simulations=100000):
    """
    Simulates a specific ping-pong game mechanism involving a ball queue, a limited-capacity platform,
    and three random pistons with complex ejection/replacement rules. Used to calculate the 
    probability of each ball number being 'ejected by a piston' (winning) versus being 'released' (eliminated).

    Args:
        num_balls (int): Total number of balls in the queue, numbered 1 to N.
        num_simulations (int): Monte Carlo iterations.

    Returns:
        dict: A dictionary containing 'win_probabilities' and 'best_choice'.
    """
    # Step 1: Validate inputs
    if not isinstance(num_balls, int) or num_balls < 1:
        raise ValueError(f"Input 'num_balls' must be a positive integer. Got: {num_balls}")
    
    if not isinstance(num_simulations, int) or num_simulations < 1:
        raise ValueError(f"Input 'num_simulations' must be an integer >= 1. Got: {num_simulations}")

    # Step 2: Initialize win_counts
    # Keys are integers 1 to N, initialized to 0
    win_counts = {i: 0 for i in range(1, num_balls + 1)}

    # Step 3: Begin simulation loop
    for _ in range(num_simulations):
        # Step 3a: Initialize queue and platform
        deck = list(range(1, num_balls + 1))
        platform = []
        
        # Initial fill of the platform (max 3 balls)
        for _ in range(3):
            if deck:
                platform.append(deck.pop(0))
            else:
                platform.append(None)
        
        # Helper function to refill balls from deck
        def get_next_balls(count):
            new_balls = []
            for _ in range(count):
                if deck:
                    new_balls.append(deck.pop(0))
                else:
                    new_balls.append(None)
            return new_balls

        # Step 4: Simulate piston firing process
        # Continue while there are still balls on the platform
        while any(ball is not None for ball in platform):
            # Select a random piston (1, 2, or 3)
            piston = random.randint(1, 3)
            
            # Current platform state
            p1, p2, p3 = platform[0], platform[1], platform[2]
            
            # Step 5: Apply Piston Rules
            if piston == 1:
                # If Piston 1: Eject Pos 1 (Win). Move Pos 2->1, Pos 3->2. Refill 1 ball.
                if p1 is not None:
                    win_counts[p1] += 1
                
                incoming = get_next_balls(1)
                platform = [p2, p3, incoming[0]]
                
            elif piston == 2:
                # If Piston 2: Eject Pos 2 (Win). Release Pos 1 (Loss). Move Pos 3->1. Refill 2 balls.
                if p2 is not None:
                    win_counts[p2] += 1
                
                # p1 is released (eliminated), so we don't increment its win count
                
                incoming = get_next_balls(2)
                platform = [p3, incoming[0], incoming[1]]
                
            elif piston == 3:
                # If Piston 3: Eject Pos 3 (Win). Release Pos 1 (Loss). Move Pos 2->1. Refill 2 balls.
                if p3 is not None:
                    win_counts[p3] += 1
                    
                # p1 is released (eliminated), so we don't increment its win count
                
                incoming = get_next_balls(2)
                platform = [p2, incoming[0], incoming[1]]

    # Step 8: Calculate probabilities
    win_probabilities = {}
    for i in range(1, num_balls + 1):
        win_probabilities[str(i)] = win_counts[i] / num_simulations

    # Find the ball with the highest probability
    best_ball_num = max(win_counts, key=win_counts.get)

    # Step 9: Format output object
    return {
        "win_probabilities": win_probabilities,
        "best_choice": best_ball_num
    }
    \end{lstlisting}
\clearpage
\appsection{Tool Implementation for \textit{Youtube Audio Transcriber}}
\begin{figure*}[!htbp]
    \centering
    \includegraphics[width=\textwidth]{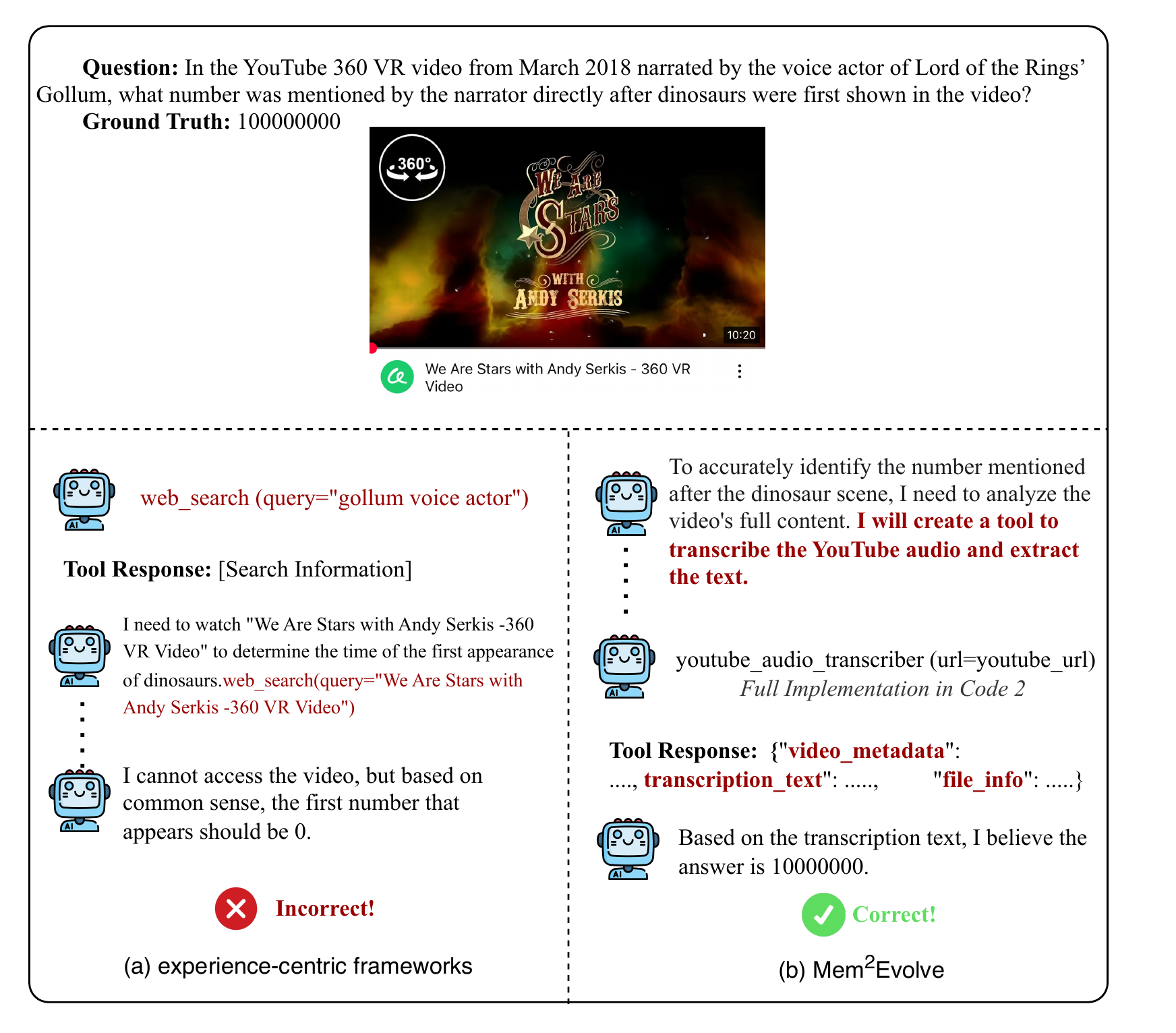}
    \caption{\textbf{Case Study 2 on YouTube Video Subtitle Extraction.} When initialized with only a web search tool, (a) experience-centric frameworks fail to handle tasks situated beyond their capability boundary, such as retrieving internal video content, leading to incorrect answers based on general common sense. In contrast, (b) Mem$^2$Evolve leverages the guidance of accumulated experience to dynamically generate high-quality tools (e.g., a custom subtitle transcriber), effectively breaking through capability boundaries to access the necessary context and derive the correct answer.}
    \label{fig:case2}
\end{figure*}

As illustrated in Code~\ref{code:youtube audio}, when the model encounters complex tasks in open-world scenarios, such as "In the YouTube 360 VR video narrated by the voice actor of Gollum from \textit{The Lord of the Rings} in March 2018, what number does the narrator directly mention after the dinosaur first appears?", it is necessary to perform semantic analysis on the audio content of YouTube videos to obtain the correct answer. To address such problems, the Agent autonomously generates the \textit{Youtube Audio Transcriber} tool, which integrates open-source libraries including yt-dlp, pytube, and speech\_recognition to enable audio transcription for arbitrary YouTube video URLs with multi-language support, thereby generalizing to similar tasks. The design of this tool demonstrates three key characteristics: First, the implementation leverages mature open-source components rather than building from scratch, significantly improving development efficiency and reliability. Second, the tool exhibits strong robustness, effectively handling diverse input formats and various edge cases. Third, the tool adheres to the principle of generality, targeting a category of tasks rather than a single problem, thus possessing good reusability to support the execution of subsequent similar tasks.

\begin{lstlisting}[style=pythonstyle, caption={Tool Implenmenation for Youtube Audio Transcriber}, label={code:youtube audio}]
import os
import math
import shutil
import json
from urllib.parse import urlparse, parse_qs
import yt_dlp
from pydub import AudioSegment
import speech_recognition as sr


def youtube_audio_transcriber(youtube_url: str, language_code: str = "en-US", chunk_duration_sec: int = 15):
    """
    Downloads audio from YouTube and transcribes it into text.
    
    Args:
        youtube_url (str): The full URL of the YouTube video.
        language_code (str): Language code for recognition (default 'en-US').
        chunk_duration_sec (int): Duration of each processing chunk in seconds (default 15).
        
    Returns:
        dict: Object containing transcription_text, video_metadata, and file_info.
    """
    
    workspace_dir = "workspace"
    temp_dir = "temp_chunks"
    
    # Ensure directories exist
    if not os.path.exists(workspace_dir):
        os.makedirs(workspace_dir)
        
    video_id = None
    downloaded_mp3_path = None
    full_transcription = ""
    video_metadata = {}
    
    try:
        # --- Step 1: Validate inputs and parse the `youtube_url` to extract the Video ID ---
        if 'youtube.com' in youtube_url:
            parsed_url = urlparse(youtube_url)
            video_id = parse_qs(parsed_url.query).get('v', [None])[0]
        elif 'youtu.be' in youtube_url:
            parsed_url = urlparse(youtube_url)
            video_id = parsed_url.path.lstrip('/')
            
        if not video_id:
            raise ValueError(f"Could not extract Video ID from URL: {youtube_url}")

        # --- Step 2: Configure `yt_dlp` options ---
        # Set output template to workspace, convert to mp3 192kbps
        output_template = os.path.join(workspace_dir, f"{video_id}_%(title)s.%(ext)s")
        
        ydl_opts = {
            'format': 'bestaudio/best',
            'outtmpl': output_template,
            'postprocessors': [{
                'key': 'FFmpegExtractAudio',
                'preferredcodec': 'mp3',
                'preferredquality': '192',
            }],
            'quiet': True,
            'no_warnings': True,
        }

        # --- Step 3: Execute the download & Extract Metadata ---
        print(f"[Tool] Starting download for Video ID: {video_id}...")
        with yt_dlp.YoutubeDL(ydl_opts) as ydl:
            info = ydl.extract_info(youtube_url, download=True)
            
            # Format duration string
            duration = info.get('duration', 0)
            m, s = divmod(duration, 60)
            h, m = divmod(m, 60)
            duration_str = f"{h:02d}:{m:02d}:{s:02d}" if h else f"{m:02d}:{s:02d}"
            
            video_metadata = {
                "title": info.get('title', 'Unknown'),
                "uploader": info.get('uploader', 'Unknown'),
                "duration_str": duration_str,
                "video_id": video_id
            }

        # --- Step 4: Locate the downloaded MP3 file ---
        # yt-dlp might replace characters in the filename, so we search by video_id
        for file in os.listdir(workspace_dir):
            if video_id in file and file.endswith('.mp3'):
                downloaded_mp3_path = os.path.join(workspace_dir, file)
                break
        
        if not downloaded_mp3_path:
            raise FileNotFoundError("Audio file not found after download process.")

        # --- Step 5: Initialize SpeechRecognition & Temp Directory ---
        recognizer = sr.Recognizer()
        
        if os.path.exists(temp_dir):
            shutil.rmtree(temp_dir)
        os.makedirs(temp_dir)

        # --- Step 6: Load MP3 & Calculate Chunks ---
        print("[Tool] Loading audio file for processing...")
        audio = AudioSegment.from_file(downloaded_mp3_path)
        
        # pydub works in milliseconds
        chunk_length_ms = chunk_duration_sec * 1000
        num_chunks = math.ceil(len(audio) / chunk_length_ms)
        
        print(f"[Tool] Audio length: {len(audio)/1000:.2f}s. Split into {num_chunks} chunks.")

        # --- Step 7, 8, 9: Iterate, Slice, Recognize, Append ---
        print("[Tool] Starting transcription...")
        
        for i in range(num_chunks):
            start_ms = i * chunk_length_ms
            end_ms = (i + 1) * chunk_length_ms
            
            # Slice audio
            chunk = audio[start_ms:end_ms]
            
            # Export to WAV (required by SpeechRecognition)
            chunk_filename = os.path.join(temp_dir, f"chunk_{i}.wav")
            chunk.export(chunk_filename, format="wav")
            
            # Recognize
            with sr.AudioFile(chunk_filename) as source:
                audio_data = recognizer.record(source)
                try:
                    text = recognizer.recognize_google(audio_data, language=language_code)
                    full_transcription += text + " "
                except sr.UnknownValueError:
                    # Audio was not understood (silence, noise, music)
                    pass
                except sr.RequestError as e:
                    print(f"[Tool] API Error on chunk {i}: {e}")
                except Exception as e:
                    print(f"[Tool] Unexpected error on chunk {i}: {e}")

        full_transcription = full_transcription.strip()

        # Get file size for report
        file_size_mb = os.path.getsize(downloaded_mp3_path) / (1024 * 1024)

        # --- Step 10: Clean up temporary files ---
        if os.path.exists(temp_dir):
            shutil.rmtree(temp_dir)

        # --- Step 11: Return result object ---
        return {
            "transcription_text": full_transcription,
            "video_metadata": video_metadata,
            "file_info": {
                "local_path": downloaded_mp3_path,
                "file_size_mb": round(file_size_mb, 2)
            }
        }

    except Exception as e:
        # Cleanup temp if error occurs
        if os.path.exists(temp_dir):
            shutil.rmtree(temp_dir)
        
        # Return error structure or raise
        return {
            "error": str(e),
            "transcription_text": "",
            "video_metadata": video_metadata if video_metadata else {},
            "file_info": {}
        }
    \end{lstlisting}
\appsection{Experience Guidance Tool Creation}

\begin{figure*}[!htbp]
    \centering
        \begin{lstlisting}[style=memory]
(*\textcolor{blue}{\#\# How to Analyze Images Using GPT-4o Multimodal Model?}*)

(*\textcolor{blue}{\#\#\# Description}*)
Parse and analyze an image file using GPT-4o multimodal model. This code can understand complex visual content, generate captions, extract tables as HTML, create SVG code for geometric shapes, and answer specific questions about images.

(*\textcolor{blue}{\#\#\# Use Cases}*)
- Product image analysis for e-commerce catalog management
- Medical image interpretation and diagnostic support
- Security and surveillance image analysis
- Educational content creation from visual materials
- Art and cultural artifact documentation
- Scientific image analysis and research documentation
- Social media content moderation and analysis

(*\textcolor{blue}{\#\#\# Tool Implementation}*)
```python
# Partial code implementation is omitted here

# Prepare API request payload
payload = {
    "model": "gpt-4o-2024-11-20",
    "messages": [
        {
            "role": "user",
            "content": [
                {
                    "type": "text",
                    "text": prompt,
                },
                {
                    "type": "image_url",
                    "image_url": {
                        "url": f"{img_type}{img_base64}"
                    }
                }
            ],
        },
    ],
    "max_tokens": 16384,
}

# Get API credentials from environment variables
api_key = os.getenv("OPENAI_API_KEY")
api_base = os.getenv("OPENAI_BASE_URL")

headers = {
    "Content-Type": "application/json",
    "Authorization": f"Bearer {api_key}"
}

# Send request to OpenAI API
response = requests.post(f"{api_base}/chat/completions", headers=headers, json=payload)

result = response.json()
output = result["choices"][0]["message"]["content"]
```
        \end{lstlisting}
\caption{\textbf{Tool Experience: Using the GPT-4o for Image Analysis.} 
This tool experience illustrates how to call the GPT-4o API to analyze images, where the agent can customize prompts to steer GPT-4o toward diverse and complex visual understanding tasks (e.g., recognition, counting, spatial reasoning, chart/diagram interpretation, and multimodal grounding). 
Each tool experience is organized into four fields: \textcolor{blue}{\textit{Title}}, \textcolor{blue}{\textit{Description}}, \textcolor{blue}{\textit{Use Cases}}, and \textcolor{blue}{\textit{Tool Implementation}}.}
\label{fig: case_tool_experience}
\end{figure*}

In this case study, we demonstrate how Mem$^2$Evolve creates a new tool under the guidance of Experience Memory. When the system already contains a memory item titled “\textbf{\textit{How to Analyze Images Using the GPT-4o Multimodal Model?}}” (Figure~\ref{fig: case_tool_experience}), the model, in generating a tool (Code~\ref{code:analyze_video}) for analyzing YouTube video content, first extracts screenshots from the video at fixed frame intervals. Subsequently, it applies GPT-4o to analyze each extracted frame and automatically constructs a complete, well-aligned prompt tailored to the analysis task.

\begin{lstlisting}[style=pythonstyle, caption={Code Implenmenation for Analyze Video for Species Tool}, label={code:analyze_video}]
import os
import json
import subprocess
import base64
import requests
import math
import shutil
import tempfile
from dotenv import load_dotenv

load_dotenv()


def analyze_video_for_species(youtube_url: str, target_subject: str = "bird species", sampling_interval: int = 10):
    """
    Analyzes a YouTube video to determine the maximum number of distinct species visible simultaneously.
    """
    
    api_key = os.getenv("OPENAI_API_KEY")
    if not api_key:
        return {"error": "Missing OPENAI_API_KEY environment variable."}

    workspace_dir = "workspace"
    if not os.path.exists(workspace_dir):
        os.makedirs(workspace_dir)

    max_species_count = 0
    best_frame_data = {
        "count": 0,
        "species_list": [],
        "description": "No data found."
    }
    best_timestamp = "00:00:00"

    try:
        # --- Step 1: Validate URL and Extract Metadata ---
        print(f"[Tool] Getting video info for: {youtube_url}")
        cmd_info = ['yt-dlp', '--dump-json', '--no-playlist', youtube_url]
        result_info = subprocess.run(cmd_info, capture_output=True, text=True, timeout=30)
        
        if result_info.returncode != 0:
            raise ValueError(f"Failed to extract video info: {result_info.stderr}")
            
        video_info = json.loads(result_info.stdout)
        duration = video_info.get('duration') # seconds
        video_id = video_info.get('id', 'unknown')
        
        if not duration:
            raise ValueError("Could not determine video duration.")

        # --- Step 2: Calculate Timestamps ---
        # Limit total checks to avoid excessive API usage cost in this demo implementation
        # For production, you might want to remove the limit or increase interval
        timestamps_sec = range(0, int(duration), sampling_interval)
        total_steps = len(timestamps_sec)
        
        print(f"[Tool] Video Duration: {duration}s. Sampling every {sampling_interval}s. Total checks: {total_steps}")

        # --- Step 3 & 4: Iterate through timestamps ---
        for idx, current_sec in enumerate(timestamps_sec):
            
            # Format timestamp HH:MM:SS
            m, s = divmod(current_sec, 60)
            h, m = divmod(m, 60)
            timestamp_str = f"{h:02d}:{m:02d}:{s:02d}"
            
            print(f"[Tool] ({idx+1}/{total_steps}) Processing timestamp: {timestamp_str}...")

            # --- Step 5: Download Segment (using ffmpeg downloader for speed) ---
            # Create a unique temp file for this segment
            temp_segment_path = os.path.join(workspace_dir, f"temp_{video_id}_{current_sec}.mp4")
            screenshot_path = os.path.join(workspace_dir, f"frame_{video_id}_{current_sec}.jpg")
            
            try:
                # Use yt-dlp with ffmpeg external downloader to fetch just a tiny snippet
                # This avoids downloading the whole video
                download_cmd = [
                    'yt-dlp',
                    '--format', 'best[height<=720]', # 720p is enough for recognition
                    '--external-downloader', 'ffmpeg',
                    '--external-downloader-args', f'ffmpeg_i:-ss {current_sec} -t 2', # download 2 seconds
                    '--output', temp_segment_path,
                    '--quiet', '--no-warnings',
                    youtube_url
                ]
                subprocess.run(download_cmd, capture_output=True, timeout=60)

                # --- Step 6: Extract Screenshot ---
                # Check if video segment exists (sometimes yt-dlp appends ext)
                found_video = None
                for ext in ['.mp4', '.webm', '.mkv']:
                    check_path = temp_segment_path.replace('.mp4', ext) # naive replacement
                    if os.path.exists(check_path): # Check exact match first if template wasn't dynamic
                        found_video = check_path
                        break
                    # Handle yt-dlp output template behavior if needed
                    if os.path.exists(temp_segment_path):
                        found_video = temp_segment_path
                        break
                
                if not found_video:
                    print(f"    -> Warning: Could not download segment for {timestamp_str}, skipping.")
                    continue

                # Capture first frame of the segment
                ffmpeg_cmd = [
                    'ffmpeg', '-i', found_video,
                    '-vframes', '1', '-q:v', '2', '-y',
                    screenshot_path
                ]
                subprocess.run(ffmpeg_cmd, capture_output=True, timeout=10)

                if not os.path.exists(screenshot_path):
                    print(f"    -> Warning: Screenshot failed for {timestamp_str}, skipping.")
                    continue

                # --- Step 7: Encode Base64 ---
                with open(screenshot_path, "rb") as image_file:
                    base64_image = base64.b64encode(image_file.read()).decode('utf-8')

                # --- Step 8: Construct AI Prompt ---
                prompt_text = (
                    f"Analyze this image specifically to count {target_subject}. "
                    f"Identify all distinct {target_subject} present. "
                    "Ignore statues, pictures, or reflections if clearly not real/live instances (unless the target is inanimate). "
                    "Return ONLY valid JSON with the following keys: "
                    "'count' (integer, number of UNIQUE species visibly distinct simultaneously), "
                    "'species_list' (array of strings, names of identified species), "
                    "'description' (string, brief summary of the scene)."
                )

                payload = {
                    "model": "gpt-4o",
                    "messages": [
                        {
                            "role": "user",
                            "content": [
                                {"type": "text", "text": prompt_text},
                                {
                                    "type": "image_url",
                                    "image_url": {"url": f"data:image/jpeg;base64,{base64_image}"}
                                }
                            ]
                        }
                    ],
                    "max_tokens": 500,
                    "response_format": {"type": "json_object"}
                }

                # --- Step 9: Call AI API ---
                headers = {
                    "Content-Type": "application/json",
                    "Authorization": f"Bearer {api_key}"
                }
                
                response = requests.post("https://api.openai.com/v1/chat/completions", headers=headers, json=payload)
                response_json = response.json()
                
                if "choices" not in response_json:
                    print(f"    -> API Error: {response_json}")
                    continue
                    
                content_str = response_json['choices'][0]['message']['content']
                analysis = json.loads(content_str)
                
                current_count = analysis.get('count', 0)
                print(f"    -> Found: {current_count} species ({analysis.get('species_list')})")

                # --- Step 10: Compare and Update ---
                if current_count > max_species_count:
                    max_species_count = current_count
                    best_timestamp = timestamp_str
                    best_frame_data = analysis
                    print(f"    -> New Record! {max_species_count} at {best_timestamp}")

            except Exception as e:
                print(f"    -> Error processing step {timestamp_str}: {e}")
            
            finally:
                # --- Step 11: Cleanup ---
                if os.path.exists(screenshot_path):
                    os.remove(screenshot_path)
                if 'found_video' in locals() and found_video and os.path.exists(found_video):
                    os.remove(found_video)

        # --- Step 12: Return Results ---
        result = {
            "max_simultaneous_species": max_species_count,
            "best_timestamp": best_timestamp,
            "identified_species": best_frame_data.get("species_list", []),
            "analysis_summary": best_frame_data.get("description", "")
        }
        
        return result

    except Exception as e:
        return {"error": str(e)}
    \end{lstlisting}
\appsection{Comparison of Tool Generation With and Without Experience Guidance}

\begin{figure*}[!htbp]
    \centering
    \includegraphics[width=\textwidth]{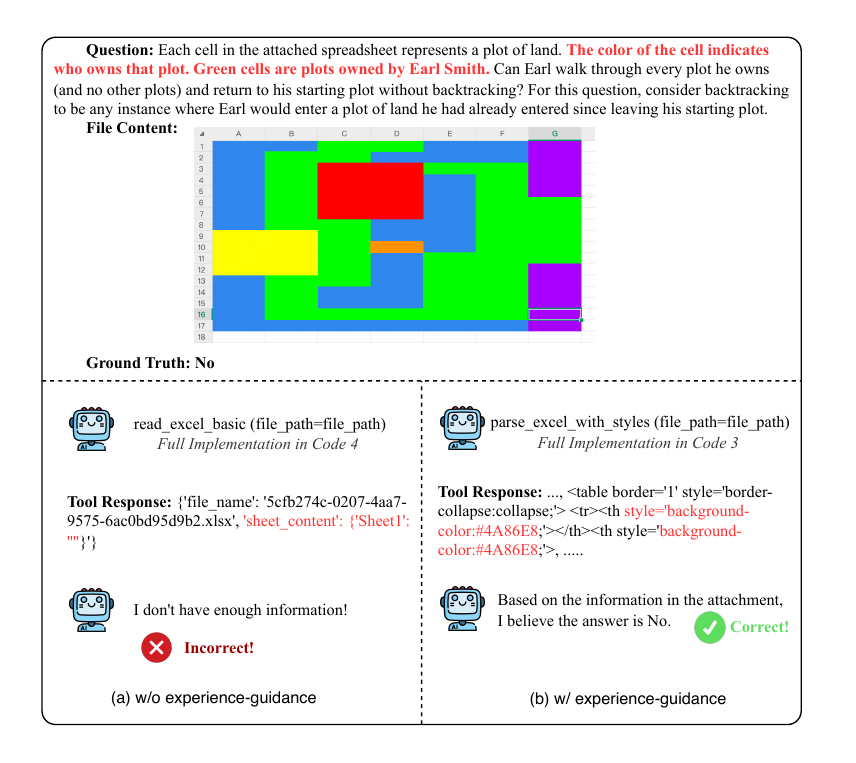}
    \caption{\textbf{Case Study 4: Experience-Guided Tool Generation for Attribute-Preserving Excel Parsing.} This case illustrates how Experience Memory guides Mem$^2$Evolve to generate task-appropriate tools that preserve critical non-textual attributes. When required to extract color-coded cells from an Excel file, Mem$^2$Evolve leverages past experience to synthesize a tool capable of accurately retrieving both cell values and their original color information in a standardized output format (full implementation in Code~\ref{code:excel with styles}). In contrast, without experiential guidance, the generated tool relies solely on pandas, which fails to retain color attributes (full implementation in Code~\ref{code:excel basic}), leading to unsuccessful task execution.}
    \label{fig:case4}
\end{figure*}

This case provides an intuitive demonstration of the guiding role of Experience Memory in the generation of new tools. Specifically, as shown in Figure~\ref{fig:case4}, Mem$^2$Evolve is required to read color-coded cells from an Excel file during task execution. With guidance from experience, the tool generated by Mem$^2$Evolve is able to accurately retrieve the target cells together with their original color information and output the results in a standardized format (full implementation in Code~\ref{code:excel with styles}). In contrast, in the absence of such experiential guidance, the model generates a tool that relies solely on pandas to read the Excel content, failing to preserve and return the color attributes (full implementation in Code~\ref{code:excel basic}). This limitation ultimately leads to unsuccessful task execution.

\begin{lstlisting}[style=pythonstyle, caption={Tool Implenmenation for Parse Excel With Styles}, label={code:excel with styles}]
import os
import pandas as pd
from openpyxl import load_workbook

def parse_excel_with_styles(file_path: str, row_limit: int = 100):
    """
    Parses an Excel or CSV file and returns the content as formatted HTML with style information preserved.
    """
    
    # Internal helper function to extract styles
    def get_cell_style(cell):
        """Extract style information from a cell and return as CSS style string."""
        styles = []

        # Check for bold formatting
        if cell.font and cell.font.bold:
            styles.append('font-weight:bold;')

        # Check for italic formatting
        if cell.font and cell.font.italic:
            styles.append('font-style:italic;')

        # Extract font color
        # Step 6 & 7: Handle Color Processing (ARGB -> RGB)
        color = getattr(cell.font, 'color', None)
        if color is not None and getattr(color, 'type', None) == 'rgb':
            rgb = getattr(color, 'rgb', None)
            if isinstance(rgb, str) and len(rgb) >= 6:
                # Slice the last 6 characters to ignore Alpha channel (ARGB -> RGB)
                styles.append(f'color:#{rgb[-6:]};')
    
        # Extract background color
        fill = getattr(cell, 'fill', None)
        fgColor = getattr(fill, 'fgColor', None)
        if fgColor is not None and getattr(fgColor, 'type', None) == 'rgb':
            rgb = getattr(fgColor, 'rgb', None)
            # Filter out transparent/invalid colors (00000000 usually means no fill in some contexts, but checking length is safer)
            if isinstance(rgb, str) and rgb != '00000000' and len(rgb) >= 6:
                styles.append(f'background-color:#{rgb[-6:]};')
        
        return ''.join(styles)

    # Step 1: Validate file existence and format
    if not os.path.exists(file_path):
        return {"error": f"Error: File '{file_path}' does not exist.", "html_content": "", "file_metadata": {}}

    supported_formats = ['.xlsx', '.xls', '.csv']
    file_ext = os.path.splitext(file_path)[1].lower()

    if file_ext not in supported_formats:
        return {"error": f"Error: Unsupported file format '{file_ext}'.", "html_content": "", "file_metadata": {}}

    html_output = ""
    metadata = {
        "file_type": file_ext,
        "sheet_names": []
    }

    try:
        # Step 2: Handle CSV files
        if file_ext == '.csv':
            df = pd.read_csv(file_path)
            metadata["sheet_names"] = ["csv_data"]
            
            html_output += f"<h2>CSV : {os.path.basename(file_path)}</h2>\n"
            html_output += f"<p>Rows: {df.shape[0]}, Columns: {df.shape[1]}</p>\n"
            html_output += "<table border='1'>\n"
            
            # Add header
            html_output += "<tr>"
            for col in df.columns:
                html_output += f"<th>{col}</th>"
            html_output += "</tr>\n"
            
            # Add data rows
            for i, row in df.head(row_limit).iterrows():
                html_output += "<tr>"
                for value in row:
                    val_str = str(value) if pd.notna(value) else ""
                    html_output += f"<td>{val_str}</td>"
                html_output += "</tr>\n"
            
            if len(df) > row_limit:
                html_output += f"<tr><td colspan='{len(df.columns)}'>... ({len(df) - row_limit} more rows)</td></tr>\n"
            
            html_output += "</table>\n"

        # Step 3: Handle Excel files
        else:
            # data_only=True is essential to get values instead of formulas
            wb = load_workbook(file_path, data_only=True)
            metadata["sheet_names"] = wb.sheetnames
            
            html_output += f"<h1>Excel: {os.path.basename(file_path)}</h1>\n"
            
            # Step 4: Iterate through sheets
            for sheet in wb.worksheets:
                html_output += f"<h2>Sheet: {sheet.title}</h2>\n"
                
                max_row = sheet.max_row
                max_col = sheet.max_column
                
                html_output += f"<p>Rows: {max_row}, Columns: {max_col}</p>\n"
                html_output += "<table border='1' style='border-collapse:collapse;'>\n"
                
                # Step 5: Process rows and cells
                # enumerate(..., 1) makes i start at 1
                for i, row in enumerate(sheet.iter_rows(max_row=min(max_row, row_limit)), 1):
                    html_output += "<tr>"
                    for cell in row:
                        tag = "th" if i == 1 else "td"  # Assume first row is header
                        
                        # Step 6: Extract style
                        style = get_cell_style(cell)
                        value = cell.value if cell.value is not None else ""
                        
                        # Step 8: Construct HTML with inline styles
                        if style:
                            html_output += f"<{tag} style='{style}'>{value}</{tag}>"
\end{lstlisting}

\bigskip

\begin{lstlisting}[style=pythonstyle, caption={Code Implenmenation for Read Excel Basic}, label={code:excel basic}]
import os
import pandas as pd

def read_excel_basic(file_path: str, preview_rows: int = 50):
    """
    Basic Excel reader using standard Pandas functionality.
    Fails to capture style information required for color-based riddles.
    """
    
    # Step 1: Validate file existence
    if not os.path.exists(file_path):
        return {"error": f"File '{file_path}' not found."}

    file_ext = os.path.splitext(file_path)[1].lower()
    data_output = {}

    try:
        # Step 2: Read file based on extension
        # Pandas read_excel defaults to reading values
        if file_ext == '.csv':
            df = pd.read_csv(file_path)
            # Convert to markdown-style string for readability
            data_output['csv_data'] = df.head(preview_rows).to_markdown(index=False)
        
        elif file_ext in ['.xlsx', '.xls']:
            # sheet_name=None reads all sheets into a dictionary
            sheets = pd.read_excel(file_path, sheet_name=None)
            
            for sheet_name, df in sheets.items():
                # Replace NaNs with empty strings for cleaner looking tables
                df_clean = df.fillna("")
                
                # We limit the rows to avoid overwhelming the context window
                preview_df = df_clean.head(preview_rows)
                
                data_output[sheet_name] = preview_df.to_markdown(index=False)
        else:
            return {"error": "Unsupported file format."}

        # Step 4: Return result
        return {
            "file_name": os.path.basename(file_path),
            "sheet_content": data_output,
            "note": "Visual styles (colors, fonts) were not extracted."
        }

    except Exception as e:
        return {"error": f"Failed to parse file: {str(e)}"}
    \end{lstlisting}

\end{document}